%% file: main.tex
\documentclass[runningheads]{llncs}

\usepackage{eccv}

\usepackage{eccvabbrv}

\usepackage{graphicx}
\usepackage{booktabs}

\usepackage[accsupp]{
    axessibility
}

\usepackage{hyperref}

\usepackage{orcidlink}

\usepackage{xspace}
\usepackage{wrapfig}
\usepackage{graphicx}
\usepackage{lipsum}
\usepackage{multirow}
\usepackage{multicol}
\usepackage[table]{xcolor}
\usepackage{booktabs}
\usepackage{array}
\usepackage{tcolorbox}

\newcommand{\dataname}{MacroData\xspace}
\newcommand{\benchname}{MacroBench\xspace}
\newcommand{\mypara}[1]{\vspace{3pt plus 1pt minus 1pt}
\noindent
\textbf{#1}}

\begin{document}
    \title{MACRO: Advancing Multi-Reference Image Generation with Structured Long-Context Data}

    \titlerunning{MACRO}

    \author{Zhekai Chen\inst{1} \and Yuqing Wang\inst{1} \and Manyuan Zhang\inst{2}$^{\dag}$ \and Xihui Liu\inst{1}$^{\dag}$}

    \authorrunning{Z. Chen et al.}

    \institute{HKU MMLab \and Meituan}

    \maketitle
    \begingroup
    \renewcommand\thefootnote{}
    \footnotetext{$^{\dag}$ Corresponding authors.}
    \endgroup

    \input{secs/0-abstract.tex}
    \input{secs/1-introduction.tex}
    \input{secs/2-related.tex}
    \input{secs/3-dataset.tex}
    \input{secs/4-benchmark.tex}
    \input{secs/5-exp.tex}
    \input{secs/6-conclusion.tex}

    \newpage

    \appendix

    \section*{Appendix}
    \input{secs/appendix/1-data}
    \input{secs/appendix/2-bench}
    \input{secs/appendix/3-exp}
    \input{secs/appendix/4-failure}
    \input{secs/appendix/5-others}

    \newpage
    \input{secs/appendix/w-data}
    \input{secs/appendix/x-prompt}

    \bibliographystyle{splncs04}
    \bibliography{main}
\end{document}

%% file: secs/0-abstract.tex
\input{figs/teaser.tex}

\begin{abstract}
    Generating images conditioned on multiple visual references is critical for real-world applications such as multi-subject composition, narrative illustration, and novel view synthesis, yet current models suffer from severe performance degradation as the number of input references grows. We identify the root cause as a fundamental data bottleneck: existing datasets are dominated by single- or few-reference pairs and lack the structured, long-context supervision needed to learn dense inter-reference dependencies. To address this, we introduce \dataname, a large-scale dataset of 400K samples, each containing up to 10 reference images, systematically organized across four complementary dimensions---Customization, Illustration, Spatial reasoning, and Temporal dynamics---to provide comprehensive coverage of the multi-reference generation space. Recognizing the concurrent absence of standardized evaluation protocols, we further propose \benchname, a benchmark of 4,000 samples that assesses generative coherence across graded task dimensions and input scales. Extensive experiments show that fine-tuning on \dataname yields substantial improvements in multi-reference generation, and ablation studies further reveal synergistic benefits of cross-task co-training and effective strategies for handling long-context complexity. The dataset and benchmark will be publicly released.

    \keywords{Multi-reference image generation \and Dataset \& Benchmark \and In-context generation.}
\end{abstract}

%% file: figs/teaser.tex
\begin{figure}[htbp]
    \vspace{-20pt}
    \centering
    \includegraphics[width=0.95\textwidth]{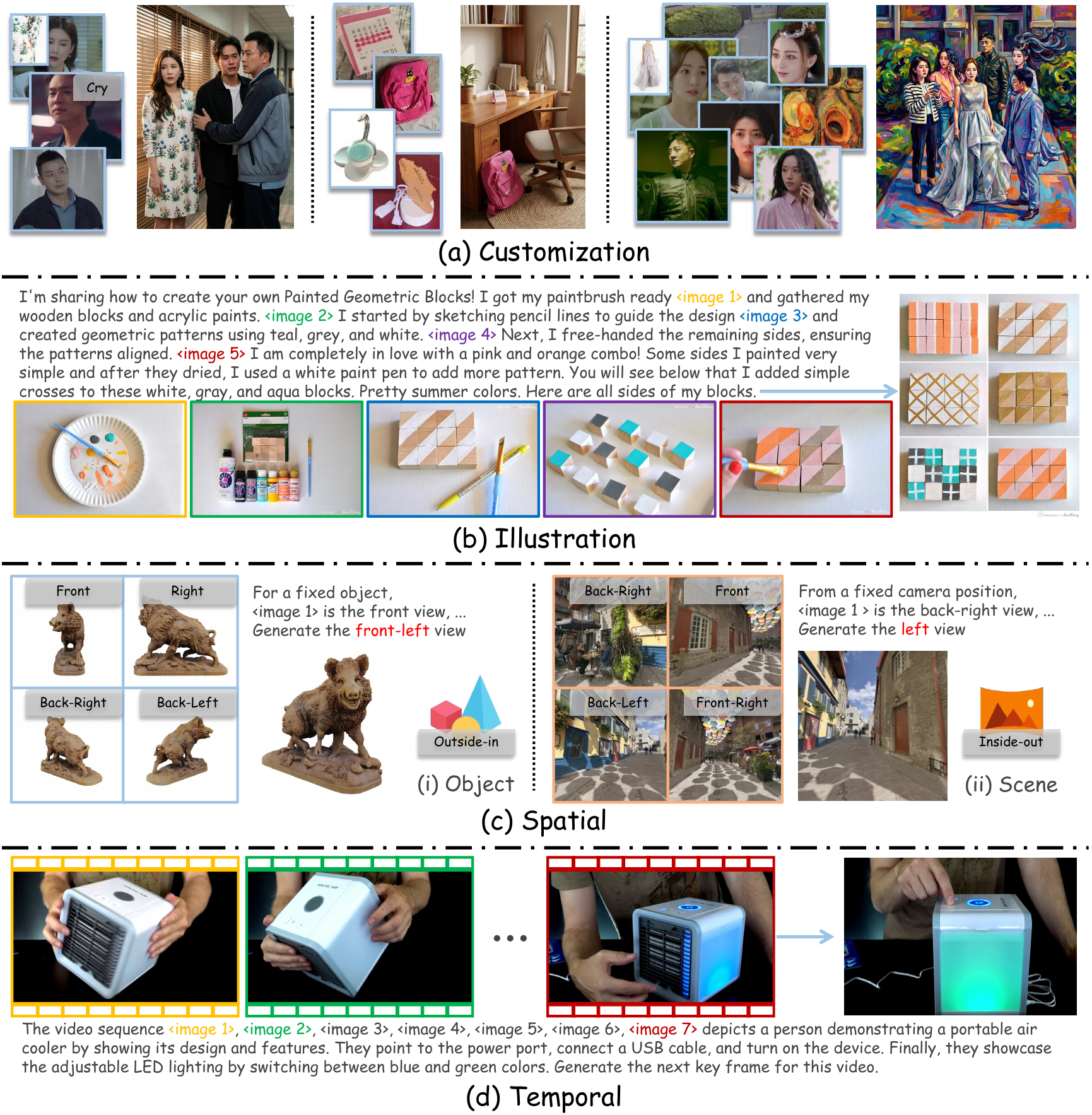}
    \vspace{-8pt}
    \caption{\textbf{Overview of \dataname.} \dataname contains 400K high-quality samples with up to 10 input images across four long-context multi-reference image generation tasks: (a) \textbf{Customization}: generating compositions conditioned on multiple reference images, (b) \textbf{Illustration}: producing illustrative images based on multimodal context, (c) \textbf{Spatial}: predicting novel view images given multiple views, specifically including outside-in objects and inside-out scenes, and (d) \textbf{Temporal}: forecasting future frames based on historical sequence. Each task is composed of 100k samples, split into different numbers of reference images, including 1-3, 4-5, 6-7, and 8-10 input images. }
    \label{fig:teaser}
    \vspace{-10pt}
\end{figure}

%% file: secs/1-introduction.tex
\section{Introduction}
\label{sec:intro}
\vspace{-5pt}

In-context generation~\cite{wu2025omnigen2,labs2025flux,sheynin2024emu,deng2025emerging,
xie2023omnicontrol,chen2025unireal} has become a prominent paradigm in image 
synthesis~\cite{rombach2022high,podell2023sdxl,chen2023pixart,blackforestlabs2024flux,
xie2024sana,lipman2022flow,esser2024scaling}, enabling models to generate images 
directly conditioned on interleaved text and visual references.
While recent advances have achieved impressive results on single- and 
few-reference tasks such as identity-preserving generation and style transfer, many 
real-world scenarios---multi-subject composition, narrative illustration, novel view 
synthesis---naturally demand reasoning over a larger set of reference images. This 
\emph{multi-reference} setting is substantially more challenging and remains largely 
unsolved.

Even the most capable open-source models struggle in this regime. For instance, OmniGen2~\cite{wu2025omnigen2} is constrained to a maximum of five 
input images, restricting its utility for more complex compositions, and 
Bagel~\cite{deng2025emerging}, despite being theoretically designed to accept unlimited inputs, exhibits severe performance degradation beyond three references. 
We attribute this deficiency primarily to a critical scarcity of high-quality, 
structured multi-reference training data. Unlike single-reference generation, which 
mainly involves extracting and reproducing features from one source, multi-reference 
generation requires models to comprehend dense inter-reference 
dependencies---temporal dynamics, spatial consistency, and identity preservation 
across inputs---and seamlessly integrate them into a coherent output. Existing 
datasets~\cite{ye2025echo,liu2025opensubject,wang2025gptedit,li2025multiedit,
ye2025imgedit}, however, are predominantly composed of single- or few-reference 
pairs and thus fail to provide the necessary training signals for learning these 
complex relationships.

To bridge this gap, we introduce \textbf{\dataname} 
(\textbf{M}ulti-image d\textbf{A}taset for \textbf{C}ontext-\textbf{R}eferencing 
generati\textbf{O}n), a large-scale dataset comprising \textbf{400K} samples with 
up to 10 reference images per sample. Rather than targeting a single capability, 
\dataname is structured around four complementary dimensions essential for 
multi-reference generation: 1) \textbf{Customization}, composing multiple referents 
into coherent original scenes; 2) \textbf{Illustration}, generating context-aware images 
that complement textual narratives; 3) \textbf{Spatial}, synthesizing novel viewpoints 
from multiple input views; and 4) \textbf{Temporal}, predicting future keyframes from 
video image sequences. To ensure logical consistency and visual fidelity, we design a robust pipeline that distills 
knowledge from advanced closed-source models and meticulously filters real-world 
corpora, rather than relying on noisy web-scraped data 
(\cref{fig:customization_pipeline,fig:illustration_pipeline,fig:spatial_pipeline,fig:temporal_pipeline}).

Beyond the training data, we observe that the lack of a standardized evaluation 
protocol has also hindered progress in multi-reference generation. Existing benchmarks 
such as OmniContext~\cite{wu2025omnigen2} are limited to customization tasks with at 
most three input images, leaving the broader setting unevaluated. We therefore propose 
\textbf{\benchname}, a comprehensive benchmark comprising 4,000 samples that evaluates 
generative coherence across all four task dimensions and graded varying input scales from 1 to 
10 images. Following recent works~\cite{ye2025imgedit,ye2025echo,wei2025mico}, we 
employ an LLM-as-Judge mechanism with task-specific criteria to rigorously assess 
context sensitivity and narrative adherence.

Equipped with \dataname and \benchname, we conduct extensive experiments by fine-tuning 
state-of-the-art open-source models (e.g., Bagel~\cite{deng2025emerging}) on our 
dataset. Results show that \dataname unlocks substantial improvements in multi-reference 
generation, significantly narrowing the gap with closed-source models. Furthermore, we provide in-depth ablation studies on data scaling and task synergies to guide future research and explore potential strategies for further enhancing multi-reference generation performance.

Our main contributions are summarized as follows:
\begin{itemize}
    \item We introduce \textbf{\dataname}, a large-scale multi-reference generation 
    dataset comprising 400K samples with up to 10 reference images, structured across 
    four complementary dimensions---Customization, Illustration, Spatial, and 
    Temporal---to facilitate inter-reference dependency learning.

    \item We propose \textbf{\benchname}, a standardized benchmark with 4,000 samples 
    that evaluates multi-reference generative coherence across graded task dimensions 
    and input scales, filling a critical evaluation void in this domain.

    \item We conduct extensive experiments demonstrating that \dataname significantly 
    enhances long-context multi-reference generation, and identify effective strategies 
    including cross-task co-training and token selection for handling complex 
    multi-reference contexts.
\end{itemize}

%% file: secs/2-related.tex
\section{Related Work}
\label{sec:related}
\vspace{-5pt}

\input{figs/statistics.tex}

\mypara{In-Context Image Generation Model.} In-context image generation requires
models to jointly comprehend preceding visual and textual inputs and synthesize coherent
images conditioned on them. To this end, diverse architectural paradigms have been
explored~\cite{ma2025janusflow,xie2024show,team2024chameleon, zhou2024transfusion,shi2024lmfusion},
with recent state-of-the-art models~\cite{deng2025emerging,wu2025omnigen2,wu2025qwen,xie2025showo2,
cui2025emu3}
achieving strong results through autoregressive, hybrid, or diffusion-based
architectures with specialized vision representations. Among them, Bagel~\cite{deng2025emerging}
introduces a Mixture-of-Transformer design that separately processes
understanding and generation tokens, while OmniGen2~\cite{wu2025omnigen2} co-trains
diffusion models with LVLM hidden states for tighter vision-language alignment.
Despite these advances, open-source models remain limited to processing at most
3--5 input images~\cite{deng2025emerging,wu2025omnigen2,wu2025qwen}, with performance
degrading sharply as reference counts increase. We attribute this limitation
largely to the absence of structured data for multi-reference scenarios,
which motivates our data-centric approach.

\mypara{In-Context Image Generation Dataset.} Constructing high-quality training
data for in-context generation remains a significant challenge. Existing
datasets are built primarily through two strategies: distillation from strong generative
models~\cite{tan2025ominicontrol,wu2025less,ye2025echo,wei2025mico, ye2025imgedit}
and retrieval from real-world corpora~\cite{xu2025withanyone,liu2025opensubject}.
For instance, Echo4o~\cite{ye2025echo} and MICo~\cite{wei2025mico} prompt closed-source
models to synthesize identity-consistent pairs, while OpenSubject~\cite{liu2025opensubject}
extracts and matches relevant images from web pages and videos. However, these
datasets are limited in two critical aspects: they focus narrowly on customization
and editing tasks, and they rarely include more than three to five reference
images per sample. This dual limitation---in both task diversity and input scale---hinders
the development of models capable of general, long-context multi-reference
generation.

\mypara{In-Context Image Generation Benchmark.} Evaluating in-context generation
poses unique challenges, as outputs must be assessed for consistency with
multiple heterogeneous inputs spanning different modalities and semantic roles. Recent
benchmarks~\cite{wei2025mico,ye2025imgedit,wu2025omnigen2, liu2025opensubject} adopt
an LLM-as-Judge paradigm following text-to-image evaluation practices~\cite{ghosh2023geneval,hu2024ella}.
For example, OmniContext~\cite{wu2025omnigen2} employs GPT-4.1~\cite{openai2024gpt4dot1}
to score prompt adherence and subject consistency. However, existing benchmarks are
restricted to customization and editing scenarios with 
at most a few input images,
lacking evaluation coverage for spatial reasoning, temporal coherence, and the
systematic scaling of input references.

\vspace{-10pt}

%% file: figs/statistics.tex
\begin{figure}[tbp]
    \centering
    \includegraphics[width=0.95\textwidth]{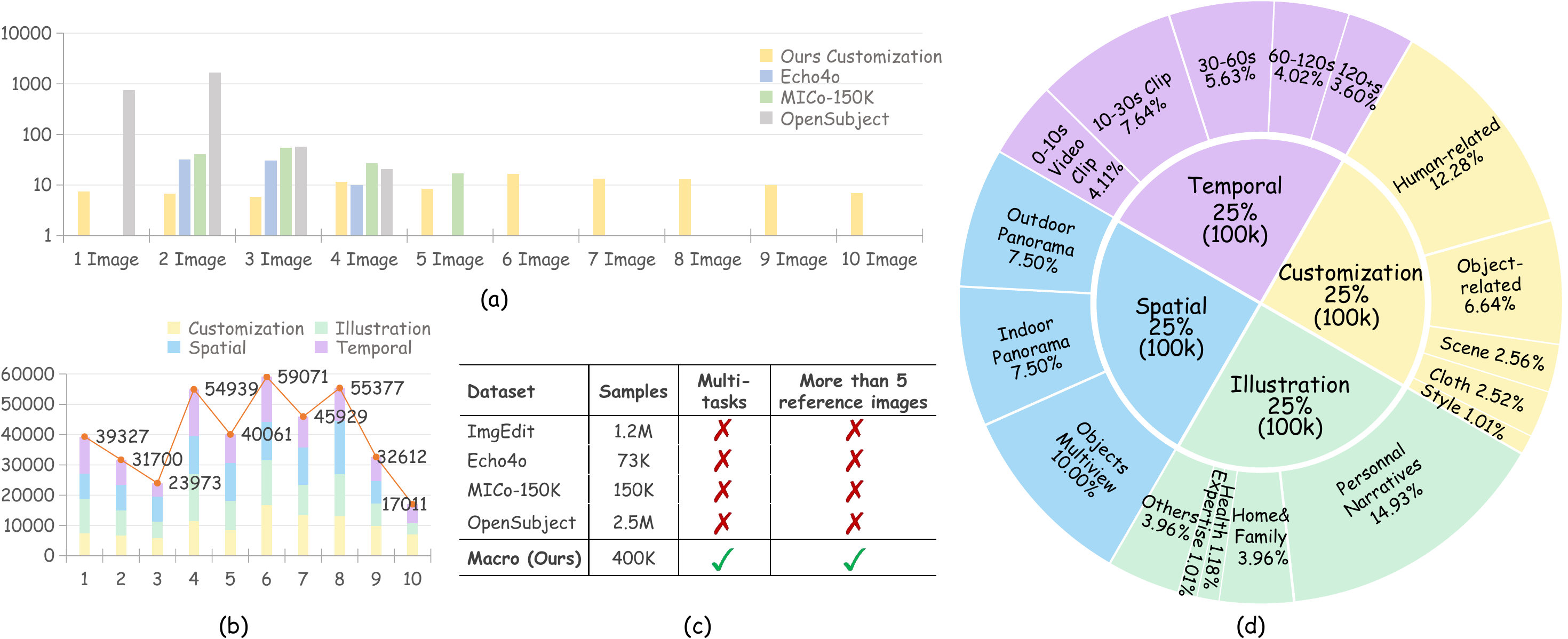}
    \vspace{-8pt}
    \caption{\textbf{Statistics of \dataname.} \dataname comprises four tasks, each containing 100k samples. (a) The number of input images in the customization subtask averaged 5.84 images per sample, with a maximum of 10. (b) The number of input images across all tasks averaged 5.44 images per sample. (c) Comparison among different datasets. (d) The distribution of data composition for each task.}
    \label{fig:statistics}
    \vspace{-18pt}
\end{figure}

%% file: secs/3-dataset.tex
\section{\dataname}
\label{sec:dataset}
\vspace{-5pt}
\subsection{Overview}
\vspace{-5pt}

\looseness=-1
We introduce \dataname, a comprehensive dataset of \textbf{400K} samples for multi-reference image generation. Addressing the scarcity of many-to-one data, it supports up to 10 inputs per sample with an average of \textbf{5.44}. As shown in Fig.~\ref{fig:statistics}(a) and (c), it significantly surpasses datasets like OpenSubject~\cite{liu2025opensubject} and Echo4o~\cite{ye2025echo}, which lack samples beyond five references. In contrast, \dataname offers substantial long-context coverage, concentrated among more than 4 references and peaking at $\sim$59K samples for 6 references as depicted in Fig.~\ref{fig:statistics}(b). This abundance of long-context samples is crucial for models to leverage extensive visual contexts.

\mypara{Task Definition.} To ensure data balance, \dataname is equally divided into four distinct tasks with 100K samples each, as detailed in Fig.~\ref{fig:statistics}(d): 1) \textit{Customization:} Covers 5 subject categories (human, object, scene, cloth, and style). 2) \textit{Illustration:} Features 100 diverse topics (e.g., narratives and health) derived from interleaved contexts. 3) \textit{Spatial:} Focuses on 3D consistency for multiview objects and panoramas. 4) \textit{Temporal:} Captures dynamics across video clips of varying durations (0 to 120+ seconds).
\vspace{-10pt}

\subsection{Customization Subset}
\input{figs/pipeline/customization.tex}
\vspace{-5pt}
\mypara{Source Collection and Preprocessing.} As depicted in~\cref{fig:customization_pipeline}, we aggregate source metadata from OpenSubject~\cite{liu2025opensubject} for human, MVImgNet~\cite{yu2023mvimgnet} for object, DL3DV~\cite{ling2024dl3dv} for scene, Vibrant Clothes Rental Dataset~\cite{vibrentClothesRental} for cloth and WikiArt~\cite{wikiart} for style. Each undergoes tailored preprocessing: we uniformly sample video keyframes for scenes, filter out clothing images containing faces via VLMs to prevent identity leakage, and categorize styles using tags to guarantee diversity. Finally, aesthetic scoring is applied to select high-quality images.

\mypara{Sampling and Generation.} We sample metadata across these categories to build diverse composition sets, using LLMs to evaluate and iteratively resample any logically or spatially unreasonable combinations. 
The valid sets are then synthesized into target images. To ensure data fidelity, we perform a VLM-based bidirectional assessment, filtering out samples where input images are not faithfully reflected in the output or where text prompts are semantically inconsistent with the generated images.

\input{figs/pipeline/illandspa}
\vspace{-8pt}
\subsection{Illustration Subset}
\vspace{-5pt}

\textbf{Anchor Image Search.} As illustrated 
in~\cref{fig:illustration_pipeline}, we leverage large-scale interleaved 
image-text sequences from web crawls of OmniCorpus-CC-210M~\cite{li2024omnicorpus} as raw source material. Since such 
data is inherently noisy, we employ VLMs to identify ``anchor images'' 
that exhibit high semantic relevance to both the accompanying text and 
preceding images. These anchors are designated as generation targets, with 
the preceding context serving as input conditions.

\mypara{Sample Reorganization.} Since the preceding contexts are often noisy and 
textually verbose, we employ strong VLMs to regenerate each sequence. 
The VLM re-evaluates semantic relevance, discards images inconsistent with 
the narrative flow, and synthesizes a concise, coherent textual context. 
Each reorganized sample is then assigned a quality score, and low-scoring 
samples are filtered out to finalize the Illustration dataset.

\vspace{-8pt}
\subsection{Spatial Subset}
\vspace{-5pt}
\textbf{Outside-in Objects.} As shown in the upper section 
of~\cref{fig:spatial_pipeline}, we construct this subtask from multi-view 
3D object renderings data from G-buffer Objaverse dataset~\cite{qiu2024richdreamer}. 
The outside-in setup focuses on capturing a central object from surrounding external viewpoints. 
To bridge the domain gap with real imagery, we 
filter out low-quality samples (e.g., transparent or white-textured 
objects) based on color saturation and brightness in HSV space. From a set 
of canonical views, we designate one as the target and randomly sample 
diverse input views from the remainder, strictly ensuring visual overlap 
for physical plausibility.

\mypara{Inside-out Scenes.} 
For inside-out scenes (\cref{fig:spatial_pipeline}, bottom), which capture the surrounding environment by looking outward from a central internal point, we start from panoramic images of DIT360~\cite{feng2025dit360}, Pano360~\cite{kocabas2021spec}, and Polyhaven~\cite{polyhaven}, 
filtering out non-standard formats (e.g., fisheye) and categorizing the 
rest into indoor and outdoor scenes using VLMs. We define canonical 
viewing directions from the panorama's center, designate a target view, 
and sample input views from the remaining directions, ensuring adequate 
spatial overlap with the target.

\vspace{-8pt}
\subsection{Temporal Subset}
\input{figs/pipeline/temporal.tex}
\vspace{-5pt}
\mypara{Video Clip Extraction.} To mitigate redundancy in raw video 
content from OmniCorpus-YT~\cite{li2024omnicorpus}, we apply shot boundary detection~\cite{soucek2024transnet} to segment videos into 
semantically distinct clips and extract the central frame of each clip as 
a representative keyframe, effectively compressing temporal information 
while preserving key visual transitions.

\mypara{Sample Construction.} We group keyframes into coherent sequences 
within scene boundaries identified via DINOv2~\cite{oquab2023dinov2} visual feature similarity 
thresholds, ensuring smooth transitions. A VLM is then applied to generate a 
descriptive summary and a quality score for each sequence, and low-scoring 
samples are discarded. The final frame of each valid sequence is 
designated as the generation target, completing the Temporal dataset.

%% file: figs/pipeline/customization.tex
\begin{figure}[tbp]
    \centering
    \includegraphics[width=0.95\textwidth]{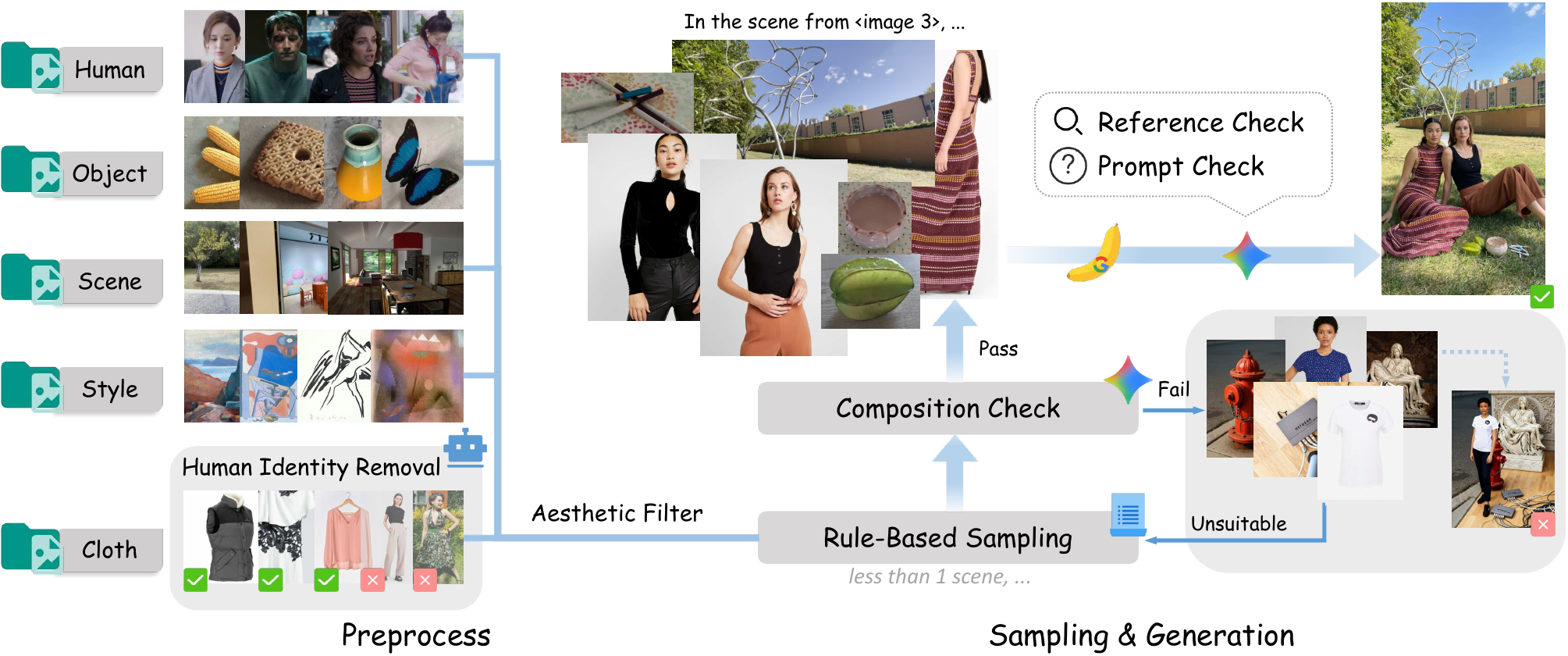}
    \vspace{-8pt}
    \caption{\textbf{Customization Subset Pipeline} composites preprocessed metadata via rule-based and VLM-reasoned sampling and applies a bidirectional assessment to ensure reference fidelity and prompt consistency.}
    \label{fig:customization_pipeline}
    \vspace{-20pt}
\end{figure}

%% file: figs/pipeline/illandspa.tex
\begin{figure}[t]
    \centering
    
    \begin{minipage}{0.49\textwidth}
        \centering
        \includegraphics[width=\linewidth]{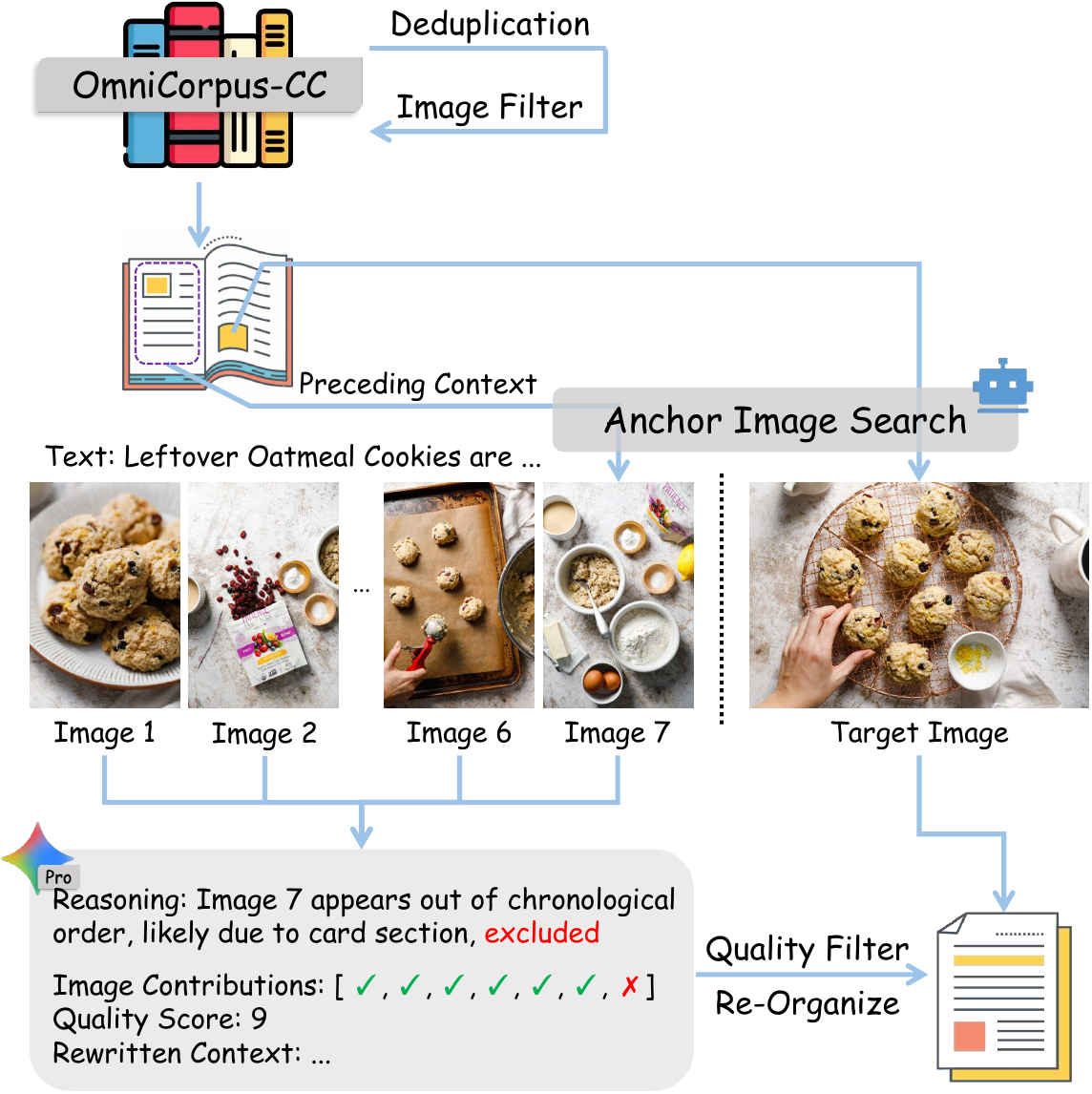}
        \vspace{-15.5pt}
        \caption{\textbf{Illustration Subset Pipeline} identifies highly relevant anchor images from interleaved data as generation targets and utilizes VLMs to rewrite and filter the preceding context for narrative coherence.}
        \label{fig:illustration_pipeline}
    \end{minipage}\hfill
    \begin{minipage}{0.48\textwidth}
        \centering
        \includegraphics[width=\linewidth]{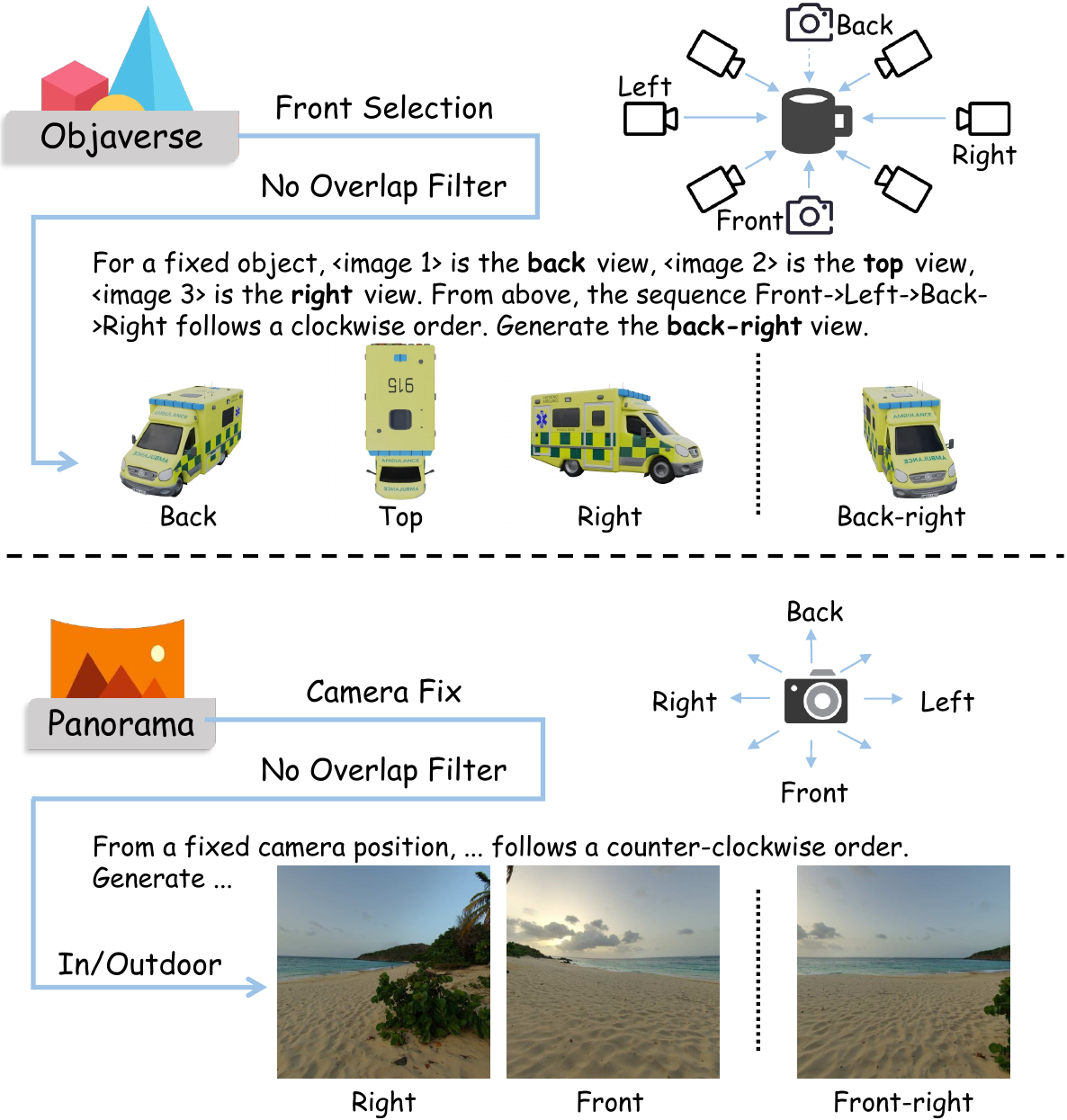}
        \vspace{-20pt}
        \caption{\textbf{Spatial Subset Pipeline} samples input and target views from canonical directions for outside-in objects and inside-out panoramas, applying spatial overlap filters to ensure plausibility.}
        \label{fig:spatial_pipeline}
    \end{minipage}
    
    \vspace{-20pt}
\end{figure}

%% file: figs/pipeline/temporal.tex
\begin{figure}[tbp]
    \centering
    \includegraphics[width=0.95\textwidth]{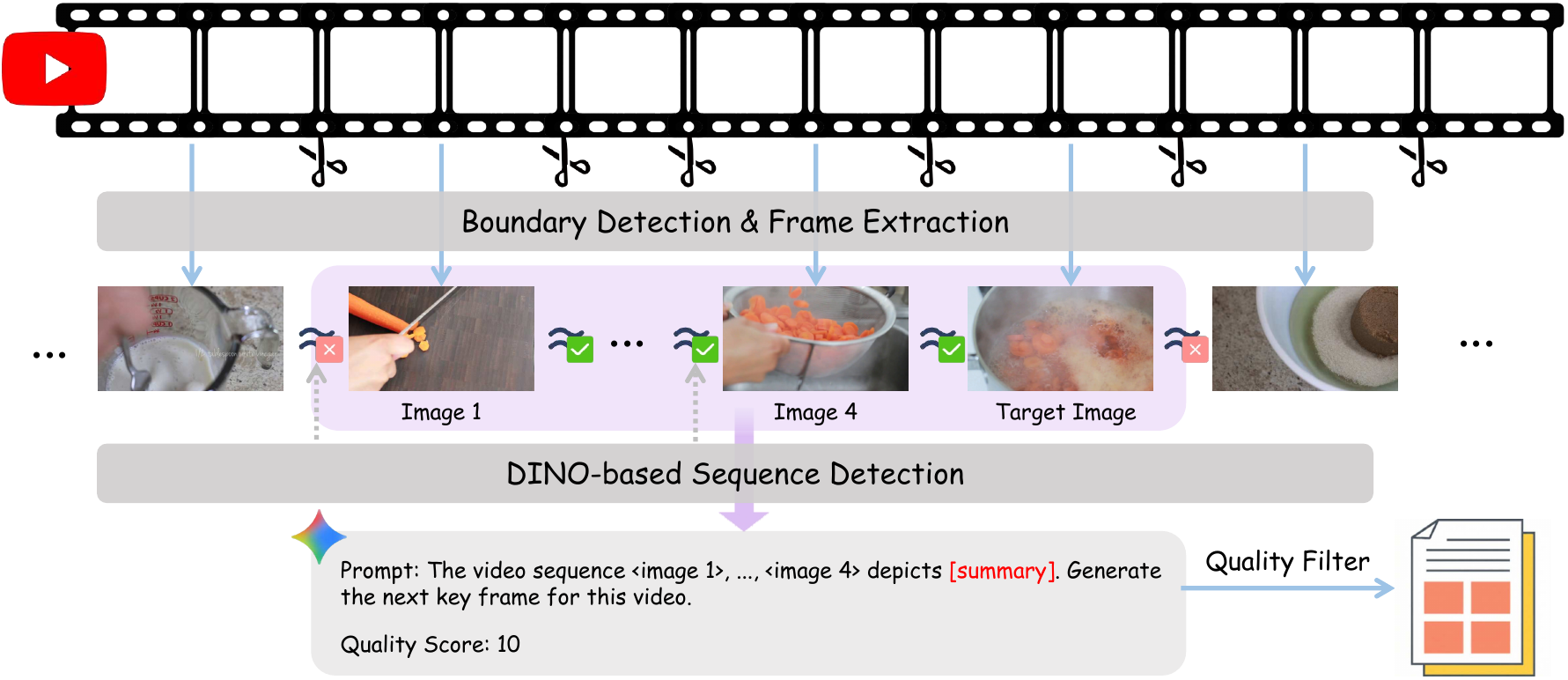}
    \vspace{-8pt}
    \caption{\textbf{Temporal Subset Pipeline} extracts video keyframes, groups them into coherent sequences using DINOv2, and utilizes a VLM to generate a summary prompt and quality score for predicting the final frame.}
    \label{fig:temporal_pipeline}
    \vspace{-20pt}
\end{figure}

%% file: secs/4-benchmark.tex
\vspace{-5pt}
\section{\benchname}
\label{sec:bench}
\vspace{-5pt}

We propose \benchname, a comprehensive benchmark evaluating in-context generation across Customization, Illustration, Spatial, and Temporal dimensions, supporting up to 10 input images.

\mypara{Tasks.} \benchname is structured along two dimensions: 1) \textbf{Task-level}: covering Customization, Illustration, Spatial, and Temporal tasks; 2) \textbf{Input-level}: categorized by input image counts (1--3, 4--5, 6--7, and 8--10). This two-dimensional 
structure enables fine-grained analysis of how model performance 
varies across both task complexity and input scale.

\mypara{Data Curation.} Following the pipeline 
in~\cref{sec:dataset}, we curate diverse evaluation sources: metadata 
(humans, objects, scenes, cloths, styles) for Customization; documents 
for Illustration; objects and panoramas for Spatial; and videos for 
Temporal. All evaluation data is strictly held out from the training 
set. We randomly sample 250 input pairs per task and image count 
category, yielding a total of 4,000 pairs,
1,000 for each task and for each image count.

\mypara{Judge Model Setting.} Following recent 
benchmarks~\cite{ghosh2023geneval,ye2025imgedit,wu2025omnigen2,
wei2025mico,liu2025opensubject}, we adopt an LLM-as-Judge evaluation 
paradigm. We find that commonly-used judge models such as 
GPT-4.1~\cite{openai2024gpt4dot1} exhibit degraded evaluation quality 
when processing multiple reference images, particularly for Spatial 
tasks requiring 3D reasoning. Through systematic comparison, we select 
Gemini-3-Flash~\cite{google2025geminiflash} as the judge model, which 
provides reliable evaluations across all task dimensions and input 
scales at reasonable cost.

We design task-specific metrics (scored 1--10) to 
evaluate distinct capabilities: \textit{Customization} assesses input 
consistency and instruction adherence via Image Consistency Score (ICS) 
and Prompt Following Score (PFS); \textit{Illustration} measures 
alignment with context text and input images via Text Consistency Score 
(TCS) and ICS; \textit{Spatial} evaluates view transformation and 
content preservation via View Transformation Score (VTS) and Content 
Consistency Score (CCS); and \textit{Temporal} examines content and 
temporal coherence via CCS and Image Sequence Consistency Score (ISCS). 
For Customization, ICS is computed per input image, and the overall ICS 
is the harmonic mean across inputs to penalize inconsistency with any 
individual reference. To ensure balanced performance across dimensions, the final score for each task is the geometric 
mean of its two metrics (e.g., $\sqrt{\text{ICS} \times \text{PFS}}$ for 
Customization), with the overall \benchname score averaged across tasks. Prompt details are provided in the Appendix.

%% file: secs/5-exp.tex
\vspace{-8pt}
\section{Experiments}
\label{sec:exp}
\vspace{-5pt}

In this section, we first validate the effectiveness of \dataname through main results 
on multi-reference generation (\cref{subsec:main_results}), then 
conduct ablation studies on dataset construction and training choices 
(\cref{subsec:ablation_study}), and finally explore advanced 
strategies for handling long-context multi-reference scenarios 
(\cref{subsec:exploration}).

\vspace{-8pt}
\subsection{Experimental Settings}
\label{subsec:exp_setting}
\vspace{-6pt}

\mypara{Baselines.} We fine-tune three open-source in-context generative models on our \dataname: Bagel~\cite{deng2025emerging}, 
OmniGen2~\cite{wu2025omnigen2}, and 
Qwen-Image-Edit-2511~\cite{wu2025qwen}. All fine-tuning runs also 
include a portion of text-to-image (T2I) data to preserve general 
generation ability. We compare against closed-source models Nano 
Banana Pro~\cite{google2025nanobanana} and 
GPT-Image-1.5~\cite{openai2025gptimage1dot5}, as well as models 
fine-tuned on alternative datasets: Echo4o~\cite{ye2025echo}, 
MICo~\cite{wei2025mico}, and OpenSubject~\cite{liu2025opensubject}. 
Note that GPT-Image-1.5 imposes a 1,000-token text limit, so only 
a subset of Illustration tasks can be evaluated; we report the 
average over successfully evaluated samples.

\mypara{Model Configuration.} We adopt a dynamic resolution strategy 
based on the number of input images to manage sequence length: 
$1024 \times 1024$ for 1--2 images, $768 \times 768$ for 3--5 
images, and $512 \times 512$ for 6--10 images, with generated images 
capped at $768 \times 768$. For Bagel~\cite{deng2025emerging}, 
ViT~\cite{dosovitskiy2020image} input tokens are further restricted 
to a maximum of $336 \times 336$.

\mypara{Benchmarks.} We evaluate on our proposed \benchname and 
OmniContext~\cite{wu2025omnigen2} for multi-reference in-context 
generation, and GenEval~\cite{ghosh2023geneval} for text-to-image 
capabilities. Open-source models generate at $768 \times 768$; 
closed-source models use their default settings.

\vspace{-8pt}
\input{tabs/maintab.tex}
\input{figs/exp/visualization.tex}
\subsection{Main Results}
\label{subsec:main_results}
\vspace{-6pt}

\mypara{\benchname.} As shown in~\cref{tab:main_table}, models 
fine-tuned on \dataname consistently outperform all open-source 
baselines across all metrics. Our fine-tuned Bagel achieves a 5.71 
average score, ranking third overall behind only the closed-source 
Nano Banana Pro and GPT-Image-1.5, and substantially improving over 
the base Bagel (3.03). Notably, it approaches Nano Banana Pro in 
Customization and surpasses it in Spatial tasks. Under identical 
architectures, \dataname also outperforms alternative fine-tuning 
datasets including Echo4o~\cite{ye2025echo}, MICo~\cite{wei2025mico}, 
and OpenSubject~\cite{liu2025opensubject}, validating its 
effectiveness.

Furthermore, while increasing input images from 1--5 to 6--10 degrades performance generally, models trained on \dataname exhibit improved robustness. For instance, our fine-tuned Qwen mitigates the severe drops in Customization (from 5.49 to 3.62) and Illustration (from 2.50 to 1.55) observed in the base model. \dataname also provides stable gains in challenging Spatial tasks where base models typically score below 1.0. More details are provided in the Appendix.

\input{tabs/omni.tex}
\mypara{OmniContext Benchmark.} To further validate the generalization capabilities of our dataset, we evaluate performance on the OmniContext benchmark~\cite{wu2025omnigen2}, which targets 1--3 image Customization tasks. Following prior studies~\cite{ye2025echo,wei2025mico,liu2025opensubject}, we fine-tune Bagel~\cite{deng2025emerging} and OmniGen2~\cite{wu2025omnigen2} on \dataname. We additionally train 
variants using only our Customization subset to isolate the effect 
of task-specific data. For baseline scores, since our configuration (\cref{subsec:exp_setting}) can improve models like OmniGen2, we report the higher score between our reproductions and literature values to ensure fair comparison.

As shown in~\cref{tab:omni_table} (full dataset in gray, Customization-only in black), \dataname achieves strong OmniContext performance despite targeting the broader multi-reference setting. It notably surpasses Echo4o~\cite{ye2025echo} (8.26 vs.\ 8.09)—a dataset purpose-built for OmniContext that also distills from closed-source models—
validating the quality of our data collection pipeline.

\mypara{Qualitative Results.} \cref{fig:visualization} presents generated images across tasks and input counts, demonstrating our method's superior capability in modeling relationships among multiple images. In Customization, it effectively integrates features from multiple images to produce coherent, contextually relevant outputs, achieving strong performance even with 10 inputs (second row). In Spatial tasks, it accurately synthesizes the target viewpoint from complex multi-view inputs. In Temporal tasks, it faithfully tracks dynamic changes across sequential frames, with strict visual consistency (e.g., the white toy sequence). These qualitative results align well with our quantitative findings, further affirming the effectiveness of \dataname on boosting multi-reference long-context image generation.

\vspace{-8pt}
\subsection{Ablation Study}
\label{subsec:ablation_study}
\vspace{-6pt}

\input{tabs/cross_table.tex}
\mypara{Cross-task Validation.} To evaluate cross-task generalization, we fine-tune four models on distinct \dataname subsets (100k samples each) and compare them to a full-dataset model across all tasks. For fair computational comparison, subset models are trained with one-quarter of the full model's iterations. As shown in~\cref{tab:cross_validation}, the full-dataset model achieves the best performance across most tasks and input counts, demonstrating the synergistic benefits of multi-task training. Moreover, subset models generally outperform the base model across all categories, indicating that each \dataname subset effectively enhances the ability to capture information and model multi-image relationships.

\input{figs/exp/data_ratio.tex}
\mypara{Data Ratio for Progressive vs.\ Non-Progressive Tasks.} We examine how the distribution of training samples across input image counts affects model performance on \textit{progressive} tasks—where task difficulty increases with input count (e.g., Customization)—and \textit{non-progressive} tasks (e.g., Temporal). We compare four sampling ratios (1:1:1:1, 2:2:3:3, 1:2:3:4, and 1:3:7:9) applied to input groups of 1--3, 4--5, 6--7, and 8--10 images. As shown in~\cref{fig:data_ratio}, upweighting large-input samples substantially boosts high-input performance on progressive tasks without hurting low-input performance, while non-progressive tasks show no such sensitivity. Based on these findings, \dataname adopts a 2:2:3:3 ratio for Customization and 1:1:1:1 for all other tasks.

\input{figs/exp/data_scaling.tex}
\mypara{Data Scaling.} We study how dataset size (1K, 5K, 10K, 20K samples) affects Customization performance, evaluated on the Customization subsets of both \benchname and OmniContext. As shown in~\cref{fig:data_scaling}, performance scales consistently with data volume, with the sharpest gains occurring between 1K and 10K. Returns diminish from 10K to 20K, suggesting approaching saturation, though larger datasets continue to stabilize training convergence. We therefore scale each task to 100K samples in \dataname.

\input{figs/exp/t2i_ratio.tex}
\mypara{Text-to-Image Tradeoff.} To analyze the trade-off between multi-reference and standard T2I generation, we evaluate models trained with varying T2I data ratios (0\%, 10\%, 20\%, 40\%) on GenEval~\cite{ghosh2023geneval} and a representative \benchname subset (50 samples per input count across tasks). As illustrated in~\cref{fig:t2i_ratio}, while T2I co-training significantly enhances GenEval performance, increasing the ratio beyond 10\% yields negligible marginal gains. Consequently, we adopt a 10\% T2I data ratio for models trained on \dataname to optimize training efficiency.

\vspace{-8pt}
\subsection{Exploration on Potential Techniques}
\label{subsec:exploration}
\vspace{-6pt}

\input{figs/selection.tex}
\mypara{Token Selection Strategies.}
Adding more input images linearly expands the token sequence, introducing long-context inefficiencies similar to those in LLMs~\cite{tang2024quest,zhang2023h2o,xiao2023efficient} and video generation~\cite{cai2025mixture,xi2025sparse,li2025radial}. 
A natural solution is to filter the context token sequence to retain only key tokens during attention calculation, referred to as token selection. Guiding the model to focus on the most informative tokens rather than the entire sequence, it effectively mitigates computational overhead and avoids being distracted by useless information. We evaluate three strategies as illustrated in~\cref{fig:selection_strategies}:
\noindent
\textit{Block-wise:} Retains top-$K$ token blocks based on mean query-key attention scores among blocks, also known as block sparse attention~\cite{yuan2025native,zhang2025spargeattention,tang2024quest}.
\noindent
\textit{Image-wise:} Selects top-$K$ images per token during the diffusion process via attention scores between each output query and the mean of image keys.
\noindent
\textit{Text-aligned:} Selects top-$K$ tokens per image during prefilling via text-image and image-image attention scores, discarding or pruning the rest~\cite{cai2025flashvlm}.

\input{tabs/exploration.tex}

\Cref{tab:selection} presents Customization results on \benchname (retention ratios for block-wise and text-aligned; image counts for image-wise)
. Image-wise selection underperforms the baseline, revealing information loss and emphasizing the necessity of cross-reference interactions. Conversely, block-wise and text-aligned strategies outperform the baseline by effectively capturing crucial multi-reference information. Text-aligned selection excels even at low retention ratios, with pruning further boosting performance. Applying selection only on VAE or ViT tokens reveals a trade-off: retaining VAE tokens is crucial given few images, whereas retaining ViT tokens becomes increasingly beneficial as image counts grow.

\mypara{Think Before Generation.} We also examine the ``think-before-generation'' strategy~\cite{guo2025deepseek}, where models generate reasoning text before the final image, as depicted in~\cref{fig:think_collage_pipeline} (a). However, this approach underperforms no-thinking baselines on multi-reference tasks as shown in~\cref{tab:think_collage_result}. We hypothesize that without explicit training for multi-reference reasoning, models inherently struggle to synthesize information across multiple images, leading to suboptimal results.

\mypara{Collage as the Proxy.} To bypass input capacity limits, a common workaround is using a collage: stitching multiple reference images into a single spatial grid as a proxy for multi-reference inputs, as displayed in~\cref{fig:think_collage_pipeline} (b). We evaluate this on Bagel~\cite{deng2025emerging} by packing inputs into one collage and appending positional descriptions ``From left to right... are <image 1> to <image n>.'' to the text prompt. Similar to the thinking strategy, collaging underperforms baselines (\cref{tab:think_collage_result}). 
We attribute this performance drop to a loss of detail, where compressing multiple images into a single collage limits the resolution of each component image.

%% file: tabs/maintab.tex
\newcolumntype{Y}{>{\centering\arraybackslash}p{1.2cm}}

\begin{table*}
  [tbp]
  \centering
  \caption{\textbf{Quantitative Comparison Results on \benchname.} ``+ X'' denotes the baseline model specifically fine-tuned on the corresponding dataset. Results of ``1--3'' and ``4--5'' are combined into ``1--5'', and ``6-7'' and ``8-10'' are combined into ``6-10''.}
  \vspace{-8pt}
  \resizebox{0.99\linewidth}{!}{
  \begin{tabular}{l|YY|YY|YY|YY|c}
    \toprule \multirow{2}{*}{\bf Model} & \multicolumn{2}{c|}{\bf Customization} & \multicolumn{2}{c|}{\bf Illustration} & \multicolumn{2}{c|}{\bf Spatial} & \multicolumn{2}{c|}{\bf Temporal} & \multirow{2}{*}{\bf Average$\uparrow$} \\
    \cmidrule(lr){2-9} & 1-5 & 6-10 & 1-5 & 6-10 & 1-5 & 6-10 & 1-5 & 6-10 & \\
    \midrule \rowcolor[HTML]{EFEFEF}\multicolumn{10}{c}{\textit{Closed-source models}} \\
    \midrule Nano Banana Pro~\cite{google2025nanobanana} & 9.41 & 7.60 & \textbf{9.12} & \textbf{8.88} & 3.28 & 3.20 & 8.21 & 7.25 & 7.12 \\
    GPT-Image-1.5~\cite{openai2025gptimage1dot5} & \textbf{9.62} & \textbf{8.76} & 8.90 & 8.84 & \textbf{3.32} & \textbf{4.24} & \textbf{8.48} & \textbf{7.84} & \textbf{7.50} \\
    \midrule \rowcolor[HTML]{EFEFEF}\multicolumn{10}{c}{\textit{Open-source models}} \\
    \midrule BAGEL~\cite{deng2025emerging} & 5.37 & 2.53 & 4.51 & 4.34 & 0.67 & 0.53 & 3.36 & 2.93 & 3.03 \\
    BAGEL + Echo4o~\cite{ye2025echo} & 6.95 & 3.98 & 4.52 & 4.14 & 0.76 & 0.80 & 3.19 & 2.41 & 3.34 \\
    BAGEL + MICo~\cite{wei2025mico} & 6.32 & 2.84 & 4.79 & 4.62 & 0.81 & 0.80 & 3.85 & 3.53 & 3.44 \\
    OmniGen2~\cite{wu2025omnigen2} & 5.21 & 2.27 & 4.72 & 3.80 & 0.67 & 1.01 & 3.00 & 2.41 & 2.89 \\
    OmniGen2 + MICo~\cite{wei2025mico} & 4.97 & 2.18 & 4.50 & 3.98 & 0.53 & 0.97 & 2.81 & 2.45 & 2.80 \\
    OmniGen2 + OpenSubject~\cite{liu2025opensubject} & 5.09 & 2.34 & 4.48 & 3.64 & 1.06 & 1.51 & 3.21 & 2.62 & 2.99 \\
    Qwen-Image-Edit-2511~\cite{wu2025qwen} & 6.41 & 0.92 & 4.69 & 2.17 & 1.57 & 0.63 & 4.02 & 1.14 & 2.69 \\
    \midrule \rowcolor{cyan!5} Bagel + Ours & \textbf{8.92} & \textbf{7.12} & \textbf{5.69} & \textbf{5.56} & \textbf{3.32} & \textbf{3.48} & \textbf{6.23} & \textbf{5.40} & \textbf{5.71} \\
    \rowcolor{cyan!5} OmniGen2 + Ours & 7.30 & 4.45 & 4.68 & 4.25 & 1.82 & 1.40 & 4.97 & 4.07 & 4.11 \\
    \rowcolor{cyan!5} Qwen + Ours & 8.31 & 4.69 & 5.04 & 3.49 & 2.76 & 2.60 & 5.03 & 2.98 & 4.36 \\
    \bottomrule
  \end{tabular}
  }
  \vspace{-18pt}
  \label{tab:main_table}
\end{table*}

%% file: figs/exp/visualization.tex
\begin{figure}[t]
    \centering
    \includegraphics[width=0.95\textwidth]{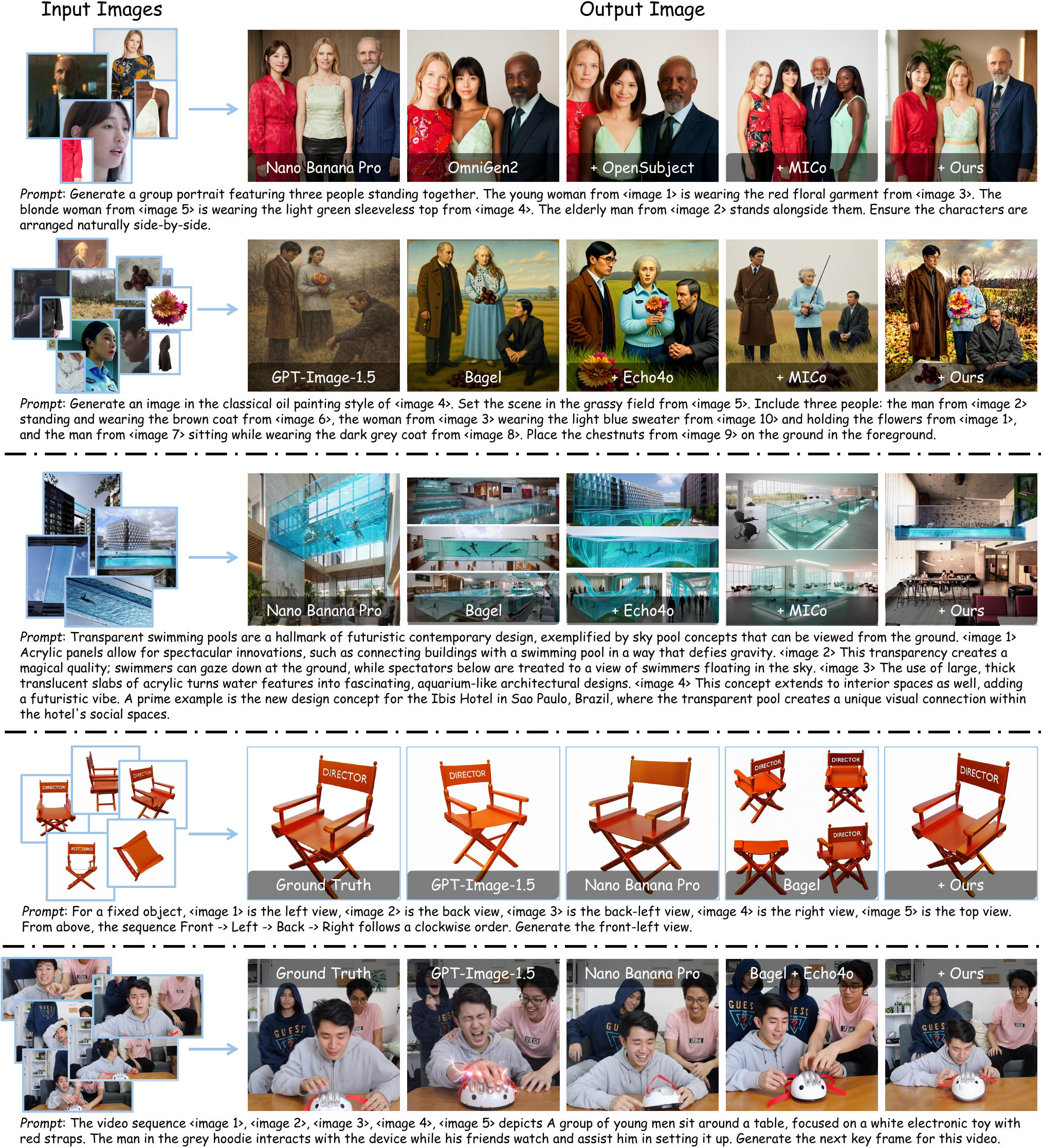}
    \vspace{-8pt}
    \caption{\textbf{Qualitative Results on All Tasks.} In each row, results of different datasets are fine-tuned from the same base model.}
    \label{fig:visualization}
    \vspace{-20pt}
\end{figure}

%% file: tabs/omni.tex
\newcommand{\rowstyle}[1]{\gdef\currentrowstyle{#1}#1\ignorespaces}
\newcolumntype{+}{>{\global\let\currentrowstyle\relax}}
\newcolumntype{^}{>{\currentrowstyle}}

\begin{table*}
   [t]
   \centering
   \caption{\textbf{Quantitative Comparison Results on OmniContext.} ``Char. + Obj.'' denotes Character + Object. Black ``+ Ours Customization'' indicates training exclusively on customization data (similar to Echo4o~\cite{ye2025echo}, MICo~\cite{wei2025mico}, and OpenSubject~\cite{liu2025opensubject}), while gray ``+ Ours All'' indicates training on the full dataset. ``$^{\dagger}$'' denotes results reported in previous works, others are reproduced under setting in~\cref{subsec:exp_setting}.}
   \vspace{-8pt}
   \resizebox{0.99\linewidth}{!}{
   \begin{tabular}{+l|^c^c|^c^c^c|^c^c^c|^c}
      \toprule \multirow{2}{*}{\bf Model} & \multicolumn{2}{c|}{\bf SINGLE} & \multicolumn{3}{c|}{\bf MULTIPLE} & \multicolumn{3}{c|}{\bf SCENE} & \multirow{2}{*}{\bf Average$\uparrow$} \\
      \cmidrule(lr){2-9} & Character & Object & Character & Object & Char. + Obj. & Character & Object & Char. + Obj. & \\
      \midrule \rowcolor[HTML]{EFEFEF}\multicolumn{10}{c}{\textit{Closed-source models}} \\
      \midrule Nano Banana Pro~\cite{google2025nanobanana} & 9.07 & 9.43 & 8.91 & 9.23 & 8.79 & 8.94 & 8.61 & 8.70 & 8.96 \\
      GPT-Image-1.5~\cite{openai2025gptimage1dot5} & \textbf{9.33} & \textbf{9.44} & \textbf{9.40} & \textbf{9.43} & \textbf{9.11} & \textbf{9.22} & \textbf{9.14} & \textbf{8.90} & \textbf{9.25} \\
      \midrule \rowcolor[HTML]{EFEFEF}\multicolumn{10}{c}{\textit{Open-source models}} \\
      \midrule BAGEL~\cite{deng2025emerging} & 7.45 & 7.40 & 5.54 & 6.56 & 7.17 & 4.59 & 5.49 & 5.95 & 6.27 \\
      BAGEL + Echo4o$^{\dagger}$~\cite{ye2025echo} & - & - & 8.07 & 7.50 & \textbf{8.29} & \textbf{8.62} & \underline{8.00} & \underline{8.08} & 8.09 \\
      BAGEL + MICo~\cite{wei2025mico} & 8.07 & 8.29 & 7.24 & 7.97 & 7.75 & 6.08 & 7.19 & 6.79 & 7.42 \\
      OmniGen2~\cite{wu2025omnigen2} & 8.22 & 8.24 & 7.45 & 7.30 & 7.78 & 7.31 & 6.47 & 7.24 & 7.50 \\
      OmniGen2 + MICo~\cite{wei2025mico} & 8.14 & 7.76 & 7.30 & 7.49 & 7.91 & 6.38 & 6.57 & 7.14 & 7.34 \\
      OmniGen2 + OpenSubject~\cite{liu2025opensubject} & 8.38 & 7.77 & 7.29 & 7.73 & 7.66 & 7.16 & 6.86 & 7.37 & 7.53 \\
      Qwen-Image-Edit-2511~\cite{wu2025qwen} & 8.49 & \underline{9.10} & 8.34 & \textbf{8.71} & 8.11 & 7.22 & 7.90 & 7.92 & 8.22 \\
      \midrule \rowcolor{cyan!5} \rowstyle{\color{black}} Bagel + Ours Customization & 8.49 & 8.76 & \textbf{8.69} & 8.17 & 8.22 & \underline{8.11} & 7.87 & 7.88 & \underline{8.26} \\
      \rowcolor{cyan!5} \rowstyle{\color{black!40}} Bagel + Ours All & 8.36 & 8.69 & 8.02 & 8.48 & 7.88 & 7.37 & 7.72 & 7.43 & 8.00 \\
      \rowcolor{cyan!5} \rowstyle{\color{black}} OmniGen2 + Ours Customization & \textbf{8.64} & 8.16 & 8.17 & 8.11 & \underline{8.28} & 8.00 & 7.62 & 8.00 & 8.12 \\
      \rowcolor{cyan!5} \rowstyle{\color{black!40}} OmniGen2 + Ours All & 8.52 & 8.51 & 7.52 & 7.94 & 8.15 & 6.59 & 6.83 & 7.98 & 7.75 \\
      \rowcolor{cyan!5} \rowstyle{\color{black!40}} Qwen + Ours All & \underline{8.60} & \textbf{9.21} & \underline{8.55} & \underline{8.69} & 8.13 & 7.75 & \textbf{8.45} & \textbf{8.19} & \textbf{8.45} \\
      \bottomrule
   \end{tabular}
   }
   \vspace{-7pt}
   \label{tab:omni_table}
\end{table*}

%% file: tabs/cross_table.tex
\begin{table*}
    [tbp]
    \centering
    \caption{\textbf{Comparison of Models Trained on Different Data Subsets.} ``+ Customization/Illustration/Spatial/Temporal'' refers to models trained exclusively on the corresponding subset, while ``+ All'' denotes the model trained on the complete dataset.}
    \vspace{-8pt}
    \resizebox{0.99\linewidth}{!}{
    \begin{tabular}{l|YY|YY|YY|YY|c}
        \toprule \multirow{2}{*}{\bf Model} & \multicolumn{2}{c|}{\bf Customization} & \multicolumn{2}{c|}{\bf Illustration} & \multicolumn{2}{c|}{\bf Spatial} & \multicolumn{2}{c|}{\bf Temporal} & \multirow{2}{*}{\bf Average$\uparrow$} \\
        \cmidrule(lr){2-9} & 1-5 & 6-10 & 1-5 & 6-10 & 1-5 & 6-10 & 1-5 & 6-10 & \\
        \midrule Bagel & 5.37 & 2.53 & 4.51 & 4.34 & 0.67 & 0.53 & 3.36 & 2.93 & 3.03 \\
        \rowcolor{cyan!5} Bagel + All & \textbf{8.92} & \textbf{7.12} & 5.69 & 5.56 & \textbf{3.32} & \textbf{3.48} & \textbf{6.23} & \textbf{5.40} & \textbf{5.71} \\
        Bagel + Customization & \cellcolor{cyan!5}8.61 & \cellcolor{cyan!5}6.43 & 5.19 & 4.92 & 0.81 & 0.64 & 4.43 & 3.17 & 4.27 \\
        Bagel + Illustration & 5.46 & 2.16 & \cellcolor{cyan!5}\textbf{5.93} & \cellcolor{cyan!5}\textbf{5.70} & 1.21 & 1.22 & 5.14 & 4.59 & 3.92 \\
        Bagel + Spatial & 5.08 & 2.24 & 4.89 & 4.78 & \cellcolor{cyan!5}2.72 & \cellcolor{cyan!5}2.98 & 4.99 & 4.13 & 3.97 \\
        Bagel + Temporal & 5.25 & 1.87 & 4.96 & 4.84 & 1.05 & 1.07 & \cellcolor{cyan!5}6.17 & \cellcolor{cyan!5}5.19 & 3.80 \\
        \bottomrule
    \end{tabular}
    } \label{tab:cross_validation}
    \vspace{-20pt}
\end{table*}

%% file: figs/exp/data_ratio.tex
\begin{figure}[tbp]
    \centering
    \includegraphics[width=0.95\textwidth]{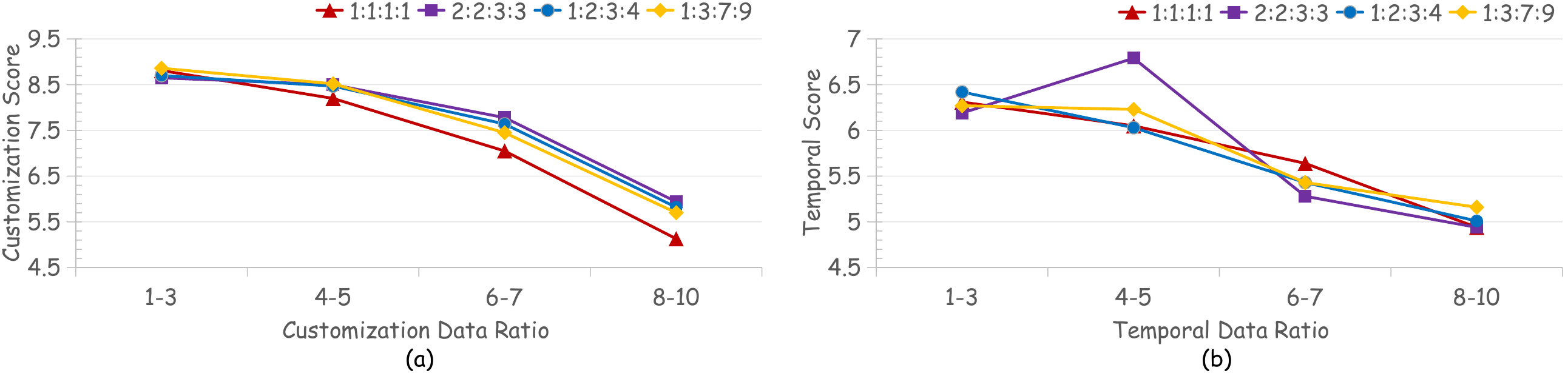}
    \vspace{-8pt}
    \caption{\textbf{Impact of Data Ratio.} (a) Performance curves on Customization tasks of our \benchname. (b) Performance curves on Temporal tasks of our \benchname. The x-axis represents the different image count categories (1--3/4--5/6--7/8--10). Each line represents a different data ratio for training data of different input counts.}
    \label{fig:data_ratio}
    \vspace{-10pt}
\end{figure}

%% file: figs/exp/data_scaling.tex
\begin{figure}[tbp]
    \centering
    \includegraphics[width=0.95\textwidth]{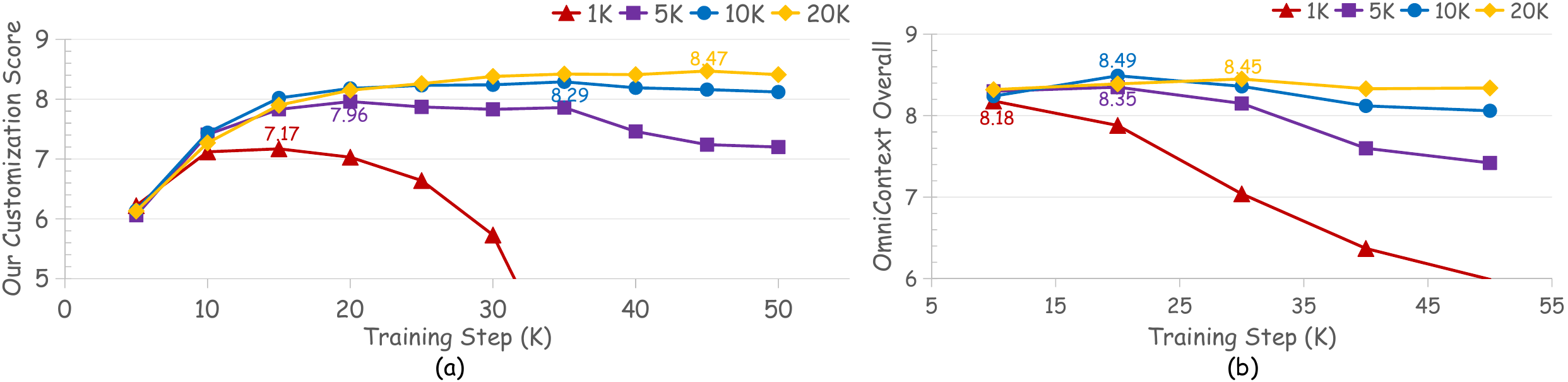}
    \vspace{-8pt}
    \caption{\textbf{Data Scaling Analysis.} (a) Performance curves on the Customization task of our \benchname. (b) Performance curves on OmniContext~\cite{wu2025omnigen2}. The x-axis represents the number of training steps. Each line represents different number of training samples.}
    \label{fig:data_scaling}
    \vspace{-20pt}
\end{figure}

%% file: figs/exp/t2i_ratio.tex
\begin{figure}[tbp]
    \centering
    \includegraphics[width=0.95\textwidth]{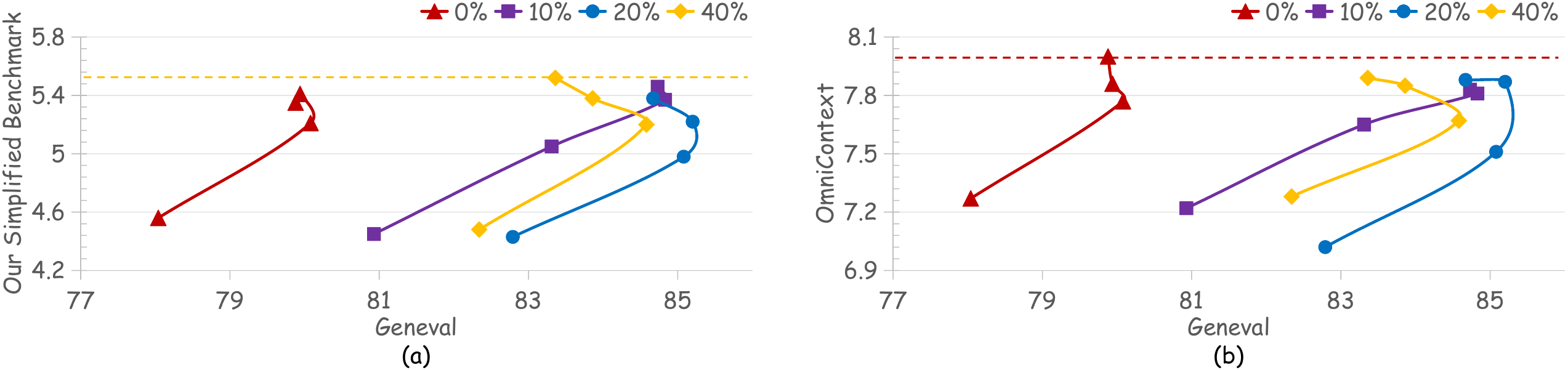}
    \vspace{-8pt}
    \caption{\textbf{Text-to-Image Data Ratio.} (a) The performance curve on the subset of our \benchname. (b) The performance curve on OmniContext~\cite{wu2025omnigen2}. The x-axis is the GenEval Score~\cite{ghosh2023geneval} for text-to-image generation evaluation. Each line represents different ratio of text-to-image data in training.}
    \label{fig:t2i_ratio}
    \vspace{-10pt}
\end{figure}

%% file: figs/selection.tex
\begin{figure}[tbp]
    \centering
    \includegraphics[width=0.95\textwidth]{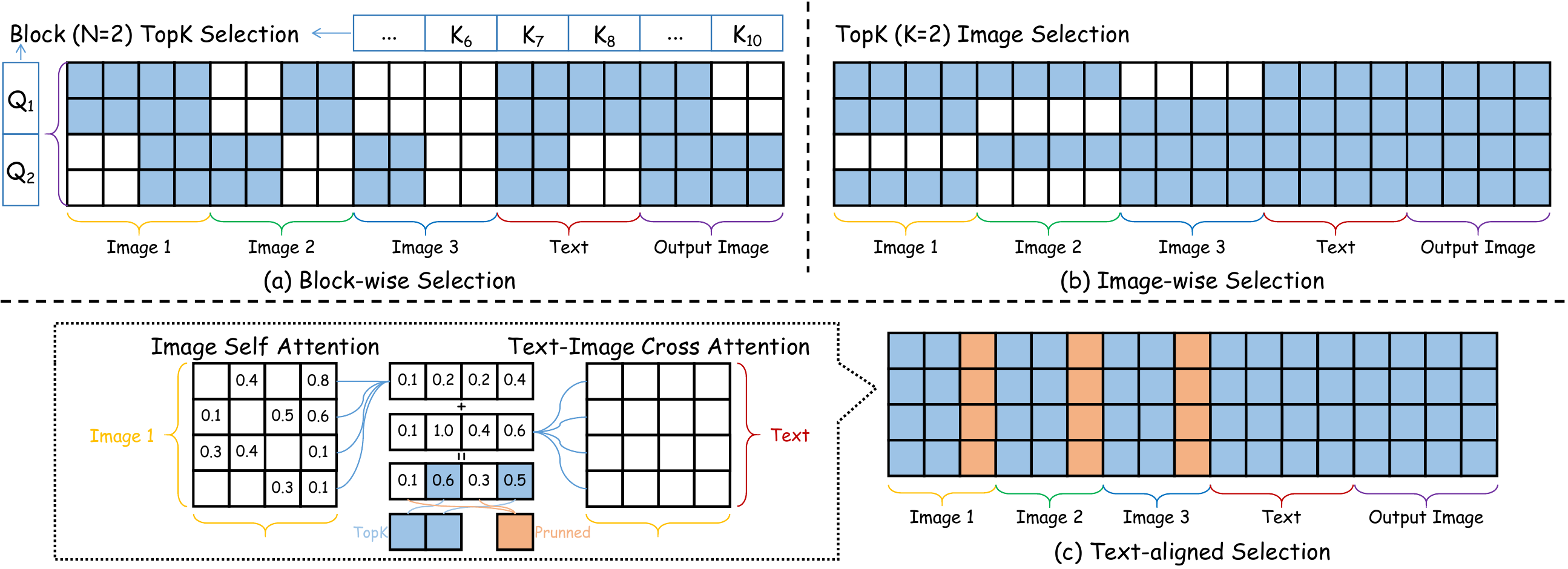}
    \vspace{-8pt}
    \caption{\textbf{Token Selection Strategies.} (a) Block-wise Selection, (b) Image-wise Selection, (c) Text-aligned Selection.}
    \label{fig:selection_strategies}
    \vspace{-22pt}
\end{figure}

%% file: tabs/exploration.tex
\begin{table*}
    [tbp]
    \centering

    \begin{minipage}[t]{0.49\linewidth}
        \centering
        \caption{\textbf{Selection Comparison.}}
        \label{tab:selection}
        \vspace{-10pt}

        \resizebox{\linewidth}{!}{
        \begin{tabular}{l|YYYY|c}
            \toprule \textbf{Settings} & \textbf{1-3} & \textbf{4-5} & \textbf{6-7} & \textbf{8-10} & \textbf{Avg$\uparrow$} \\
            \midrule Bagel + Ours & 9.00 & 8.84 & 8.01 & 6.23 & 8.02 \\
            \midrule \rowcolor[HTML]{EFEFEF}\multicolumn{6}{c}{\textit{Block-wise}} \\
            \midrule 50\% & 7.23 & 8.49 & 7.52 & 5.99 & 7.31 \\
            80\% & 8.92 & 8.88 & 8.05 & 6.60 & \underline{8.11} \\
            \rowcolor{cyan!5} 90\% & 9.04 & 9.06 & 8.19 & 6.54 & \textbf{8.21} \\
            \midrule \rowcolor[HTML]{EFEFEF}\multicolumn{6}{c}{\textit{Image-wise}} \\
            \midrule 30\% Image & 8.20 & 7.99 & 6.83 & 5.49 & 7.13 \\
            50\% Image & 8.32 & 7.95 & 7.40 & 5.99 & 7.42 \\
            \rowcolor{cyan!5} \> w/ only VAE & 8.63 & 8.39 & 7.62 & 6.04 & \textbf{7.67} \\
            \> w/ only ViT & 8.98 & 8.43 & 7.25 & 5.40 & \underline{7.52} \\
            \midrule \rowcolor[HTML]{EFEFEF}\multicolumn{6}{c}{\textit{Text-aligned}} \\
            \midrule 80\% & 9.02 & 8.78 & 7.94 & 6.29 & 8.01 \\
            50\% & 9.13 & 9.04 & 8.00 & 6.36 & 8.13 \\
            30\% & 9.03 & 9.02 & 7.90 & 6.48 & 8.11 \\
            \rowcolor{cyan!5} \> w/ only VAE & 9.04 & 8.84 & 8.01 & 6.82 & \textbf{8.18} \\
            \> w/ only ViT & 9.11 & 8.89 & 8.13 & 6.54 & \underline{8.17} \\
            \> w/ 10\% pruning & 9.02 & 9.02 & 8.20 & 6.39 & 8.16 \\
            \bottomrule
        \end{tabular}
        }
    \end{minipage}
    \hfill
    \begin{minipage}[t]{0.49\linewidth}
        \centering

        \caption{\textbf{Think \& Collage Results.}}
        \label{tab:think_collage_result}
        \vspace{-10pt}

        \resizebox{\linewidth}{!}{
        \begin{tabular}{l|YYYY|c}
            \toprule \textbf{Model} & \textbf{1-3} & \textbf{4-5} & \textbf{6-7} & \textbf{8-10} & \textbf{Avg$\uparrow$} \\
            \midrule Bagel & 6.78 & 3.96 & 2.87 & 2.19 & 3.95 \\
            Bagel + Ours & 9.00 & 8.84 & 8.01 & 6.23 & 8.02 \\
            \midrule \rowcolor[HTML]{EFEFEF}\multicolumn{6}{c}{\textit{w/ Think}} \\
            \midrule Bagel & 6.66 & 4.20 & 2.72 & 1.98 & 3.89 \\
            Bagel + Ours & 6.04 & 4.50 & 3.10 & 2.40 & 4.01 \\
            \midrule \rowcolor[HTML]{EFEFEF}\multicolumn{6}{c}{\textit{w/ Collage}} \\
            \midrule Bagel & 5.84 & 2.79 & 2.76 & 2.08 & 3.37 \\
            Bagel + Ours & 7.73 & 6.11 & 4.96 & 3.59 & 5.60 \\
            \bottomrule
        \end{tabular}
        }

        \includegraphics[width=\linewidth]{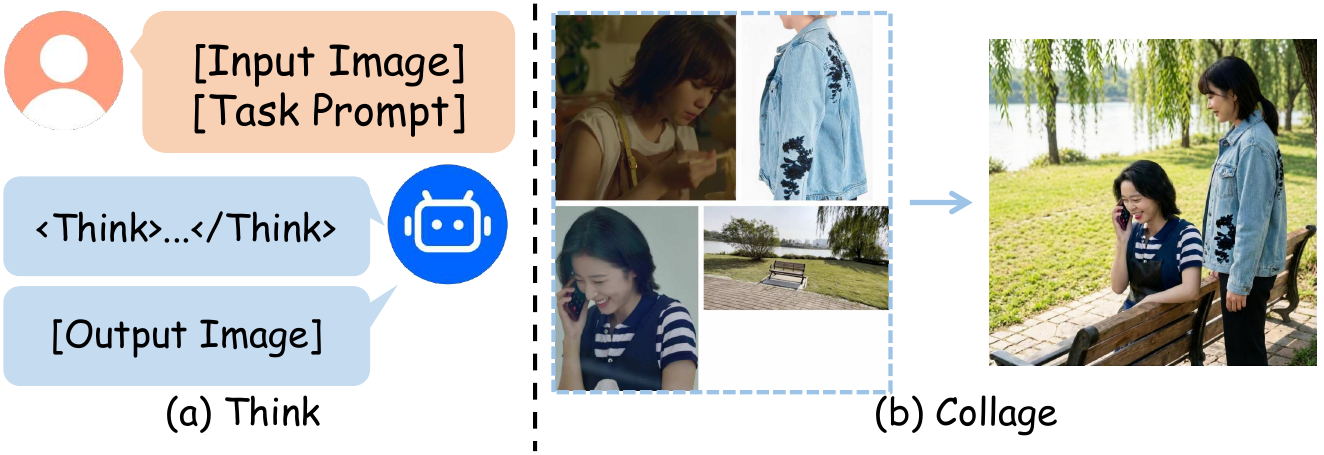}

        \vspace{-1pt}
        \captionof{figure}{\textbf{Think \& Collage Method.}} \label{fig:think_collage_pipeline}
    \end{minipage}
    \vspace{-30pt}
\end{table*}

%% file: secs/6-conclusion.tex
\vspace{-8pt}
\section{Conclusion}
\vspace{-6pt}

In this paper, we address the challenge of multi-reference image generation, where models must reason over many visual references simultaneously to produce coherent outputs. To tackle the scarcity of structured training data and the lack of standardized evaluation in this setting, we introduce \dataname, a 400K-sample dataset with up to 10 reference images spanning four complementary dimensions—Customization, Illustration, Spatial, and Temporal—and \benchname, a benchmark that evaluates generative coherence across both task types and input scales. Extensive experiments show training on \dataname yields consistent improvements in long-context multi-reference generation, and ablation studies provide practical guidelines on data construction, task synergy, and long-context efficiency. We hope \dataname and \benchname offer a useful basis for future research on in-context generation with complex multi-reference inputs.

%% file: secs/appendix/1-data.tex
\section{Detailed Data Construction Pipeline}
\label{app_sec:appendix_dataset}

\subsection{Customization Subset}
For the source collection, we utilize datasets of immense scale to ensure diversity. Specifically, the metadata encompasses over 2 million identities from OpenSubject~\cite{liu2025opensubject}, 200,000 object videos classified into 238 categories from MVImgNet~\cite{yu2023mvimgnet}, 10,000 scene videos from DL3DV~\cite{ling2024dl3dv}, 50,000 images from the Vibrant Clothes Rental Dataset~\cite{vibrentClothesRental}, and 10,000 artworks from WikiArt~\cite{wikiart}. Furthermore, the human, object, scene, and cloth categories are augmented with data from Echo4o~\cite{ye2025echo}. During preprocessing, keyframes for DL3DV~\cite{ling2024dl3dv} are extracted by uniformly sampling five frames per video. To prevent identity leakage in clothing data, we explicitly employ Qwen-3VL-8B~\cite{bai2025qwen3} to filter out images containing human faces. For style data from WikiArt, we categorize images using fine-grained ``artist-genre-style'' tags and restrict selection to a maximum of three images per category. This rigorous pipeline yields a finalized source dataset of 50,000 humans, 50,000 objects, 32,500 scenes, 30,000 cloths, and 29,292 style samples. 

During the composition phase, metadata is mixed using a strict sampling ratio of 8:6:3:2:1 across the human, object, scene, cloth, and style categories. We employ Gemini-3-Flash~\cite{google2025geminiflash} to evaluate combinations, resampling if a set is deemed unreasonable. Valid sets are then processed by Nano Banana Pro~\cite{google2025nanobanana} for target image generation. Gemini-3-Flash~\cite{google2025geminiflash} is subsequently utilized to conduct a bidirectional consistency assessment between inputs, prompts, and generated outputs. The final curated 100,000 samples for this task are distributed as 20,000, 20,000, 30,000, and 30,000 across the 1--3, 4--5, 6--7, and 8--10 image number categories, respectively.

\subsection{Illustration Subset}
The raw source material from OmniCorpus-CC-210M~\cite{li2024omnicorpus} contains an overwhelming 210 million interleaved image-text sequences. To efficiently navigate this massive scale and identify optimal ``anchor images,'' we utilize the Qwen-3VL-8B~\cite{bai2025qwen3} model. Following random sampling of candidate targets and their preceding contexts, we deploy Gemini-3-Pro~\cite{google2025gemninipro} to execute the sample reorganization phase. This model is explicitly instructed to re-evaluate semantic relevance, synthesize a concise textual context, and assign a final quality score for filtering. The resulting 100,000 high-quality illustration samples are perfectly balanced, with exactly 25,000 samples allocated to each of the four image number categories.

\subsection{Spatial Subset}
For the outside-in object subtask utilizing the 10-category G-buffer Objaverse dataset~\cite{qiu2024richdreamer}, we leverage a highly structured multi-view rendering setup. The source provides 24 views within an elevation range of $5^{\circ}$ to $30^{\circ}$ (at $15^{\circ}$ intervals), 12 views between $-5^{\circ}$ and $5^{\circ}$ (at $30^{\circ}$ intervals), alongside single top and bottom views. We define a precise canonical set of 10 views: top, bottom, left, right, front, back, and four diagonal perspectives. In this outside-in configuration, the horizontal sequence of front, left, back, and right is arranged in a clockwise order. Input views are meticulously drawn from either the $15^{\circ}$ or corresponding $30^{\circ}$ rotation sets to guarantee visual overlap. This yields 40,000 samples for the outside-in subtask, with 10,000 samples per image number category.

For the inside-out scene subtask, after filtering non-standard panoramic formats, we explicitly utilize Qwen-3VL-8B~\cite{bai2025qwen3} to classify the valid panoramas into indoor and outdoor categories. Employing the same 10 canonical view definitions as the object subtask, the inside-out horizontal sequence is instead arranged in a counter-clockwise order. Camera initialization involves a random yaw angle coupled with a pitch strictly bounded between $-10^{\circ}$ and $10^{\circ}$. This pipeline generates 30,000 indoor and 30,000 outdoor samples, with each subset stratified into 7,500 samples per image number category.

\subsection{Temporal Subset}
The raw temporal data originates from OmniCorpus-YT~\cite{li2024omnicorpus}, which comprises 10 million YouTube videos. Due to the sheer volume, we sample a representative subset of 1 million video links for raw downloading before applying TransNetv2~\cite{soucek2024transnet} for clip segmentation and central keyframe extraction. Following shot boundary identification via DINOv2~\cite{oquab2023dinov2}, we prompt Gemini-3-Flash~\cite{google2025geminiflash} to act as the evaluator, generating a descriptive summary for each visually continuous sequence and assigning the requisite quality score for filtering. The final target designates the last image of each valid sequence, resulting in 100,000 temporal samples evenly stratified with 25,000 samples per image number category.

\subsection{More Visualizations of \dataname}

We present additional data samples in~\cref{fig:app_data_cus,fig:app_data_ill,fig:app_data_spa,fig:app_data_tmp}, which separately illustrate different tasks with varying numbers of input references in our \dataname. As shown, the Customization subset encompasses diverse types of inputs, while the Illustration subset covers a wide range of topics. The Spatial subset comprises three primary categories: objects, indoor panorama, and outdoor panorama, each composed of different target viewpoints for prediction. The Temporal subset demonstrates examples with varying sequence lengths.

%% file: secs/appendix/2-bench.tex
\section{Benchmark}
\label{app_sec:benchmark}

\subsection{Benchmark Prompt}
\label{app_subsec:prompt}

As shown in~\cref{fig:prompt_customization,fig:prompt_illustration,fig:prompt_spatial,fig:prompt_temporal}, we provide the complete prompts used for each task in \benchname.
For every query, the judge model receives: (1) the task-specific
instruction, (2) all input reference images, and (3) the
generated output image.
For Spatial and Temporal tasks, the ground-truth target image is
additionally provided as an explicit reference to facilitate
fine-grained consistency assessment.

\subsection{Calculation Details}
\label{app_subsec:calc}

\subsubsection{Single-score Aggregation.}
All raw metric scores lie in $[0, 10]$.
For each sample, the two task-specific metric scores
$(M_1, M_2)$ are aggregated into a single scalar via the
geometric mean:
\begin{equation}
  S = \sqrt{M_1 \times M_2}.
  \label{eq:geomean}
\end{equation}
The geometric mean is chosen over the arithmetic mean because it
penalizes severe failure on either dimension more aggressively:
if one metric collapses to zero, the overall score also collapses,
regardless of how high the other metric is.

The metric pairs for each task are:
\begin{itemize}
  \item \textbf{Customization:}
        $S = \sqrt{\mathrm{ICS} \times \mathrm{PFS}}$,
        where ICS is the harmonic mean of per-reference
        Image Consistency Scores.
  \item \textbf{Illustration:}
        $S = \sqrt{\mathrm{TCS} \times \mathrm{ICS}}$.
  \item \textbf{Spatial:}
        $S = \sqrt{\mathrm{VTS} \times \mathrm{CCS}}$.
  \item \textbf{Temporal:}
        $S = \sqrt{\mathrm{CCS} \times \mathrm{ISCS}}$.
\end{itemize}

\subsubsection{Harmonic Mean for Customization ICS.}
Because Customization requires faithfully reproducing every
individual reference subject, ICS is first computed for each
reference image and then aggregated via the harmonic mean:
\begin{equation}
  \mathrm{ICS} = \frac{n}{\displaystyle\sum_{i=1}^{n}
  \frac{1}{\mathrm{ICS}_i}},
\end{equation}
where $n$ is the number of input reference images.
The harmonic mean ensures that low fidelity to \emph{any single}
reference significantly suppresses the overall ICS.

\subsubsection{Overall \benchname Score.}
The overall \benchname score for a model is the arithmetic mean
of its per-task scores across the four tasks.

\subsection{Validation Consistency}
\label{app_subsec:valid}

To validate the reliability of Gemini-3-Flash~\cite{google2025geminiflash} as the
judge model, we conduct a human study and measure
judge--human score consistency.

\subsubsection{Sample Construction.}
We construct a validation set of 280 samples
comprising two types:

\textit{Model-generated (non-GT) samples.}
For each combination of task $t \in$
\{Customization, Illustration, Spatial, Temporal\}
and image-count category $c \in$
\{1--3, 4--5, 6--7, 8--10\},
we randomly sample 5 outputs from each of two
representative models (Nano Banana Pro~\cite{google2025nanobanana} and
Bagel~\cite{deng2025emerging}), using a fixed random seed
($\texttt{seed}=42$).
This yields $2 \times 4 \times 4 \times 5 = 160$
non-GT samples.
Each non-GT sample is independently scored by professional
human annotators using the same task-specific rubric as the
judge model (each metric on a 1--10 scale), and the final
human score is computed with \cref{eq:geomean}.

\textit{Ground-truth (GT) samples.}
For tasks that admit well-defined reference outputs---namely
Illustration, Spatial, and Temporal---we additionally sample
10 GT items from each task and each image count, yielding
$3 \times 4 \times 10 = 120$ GT samples.
For each GT sample, the ground-truth target image is used as the
``generated'' output, and its annotation score is set to
10. Customization is excluded because no GT samples
exist for a given instruction and reference images.

\subsubsection{Correlation Metrics.}
We measure judge--human agreement via
Pearson correlation ($r$)~\cite{pearson1895vii}, 
Spearman rank correlation ($\rho$)~\cite{spearman1961proof}, 
and Kendall rank correlation ($\tau$)~\cite{kendall1938new}.
We report results under two settings: 1) \textit{Overall:} all 280 samples (GT human scores fixed at 10; non-GT human scores from annotators). 2) \textit{Human:} the 160 non-GT samples only, directly comparing judge scores with human annotator scores.

\subsubsection{Results.}
\Cref{tab:corr_overall} presents the correlation metrics across the two settings. 
Gemini-3-Flash~\cite{google2025geminiflash} achieves substantially higher agreement with
human judgments than GPT-4.1~\cite{openai2024gpt4dot1} in both the Overall and Human settings in all metrics, especially achieving a Pearson correlation of 0.821. 
This strong and consistent correlation shows its ability to correctly rate generated results, confirming our choice of
Gemini-3-Flash as the judge model.

\begin{table}[t]
  \centering
  \caption{%
    Judge--human agreement measured by Pearson ($r$), Spearman ($\rho$), and Kendall ($\tau$) correlations.
    \textbf{Overall}: all 280 samples (GT human scores assigned 10).
    \textbf{Human}: non-GT annotated samples only (160 total).
    Best per column in \textbf{bold}.%
  }
  \label{tab:corr_overall}
  \setlength{\tabcolsep}{8pt}
  \renewcommand{\arraystretch}{1.05}
  \begin{tabular}{llccc}
    \toprule
    \textbf{Judge} & \textbf{Setting} & \textbf{Pearson} & \textbf{Spearman} & \textbf{Kendall} \\
    \midrule
    \multirow{2}{*}{Gemini-3-Flash~\cite{google2025geminiflash}}
      & Overall
      & \textbf{0.821}
      & \textbf{0.795}
      & \textbf{0.665} \\
      & Human
      & \textbf{0.770}
      & \textbf{0.798}
      & \textbf{0.613} \\
    \midrule
    \multirow{2}{*}{GPT-4.1~\cite{openai2024gpt4dot1}}
      & Overall
      & 0.555
      & 0.598
      & 0.470 \\
      & Human
      & 0.447
      & 0.510
      & 0.354 \\
    \bottomrule
  \end{tabular}
\end{table}

%% file: secs/appendix/3-exp.tex
\section{Experiments Details}

\subsection{Training Settings}

Besides the dynamic resolution strategy stated in the main part, we display detailed hyperparameters for model training.

\subsubsection{BAGEL~\cite{deng2025emerging}.}
This model is fine-tuned using Fully Sharded Data Parallel (FSDP) across 32 NVIDIA H800 GPUs (4 nodes) with a frozen ViT encoder. Training utilizes a token-based dynamic batching strategy with a maximum of 32,768 tokens and a learning rate of $2 \times 10^{-5}$.

\subsubsection{OmniGen2~\cite{wu2025omnigen2}.}
This model is fine-tuned using DeepSpeed across 32 NVIDIA H800 GPUs (4 nodes). Training runs with a global batch size of 64 and a learning rate of $8 \times 10^{-7}$. The baseline of OmniGen2~\cite{wu2025omnigen2} adopts a fixed image embedding for max 5 images. To support more than 5 input references, we extend the image embedding and randomly initialize them according to a normal distribution.

\subsubsection{Qwen-Image-Edit-2511~\cite{wu2025qwen}.}
We fine-tune the DiT component using DeepSpeed across 128 NVIDIA H800 GPUs (16 nodes). Training adapts a learning rate of $1 \times 10^{-5}$.

\subsection{Evaluation Settings}

Apart from using a dynamic resolution strategy to manage context length for varying numbers of input references, we generally follow the default inference settings provided by the baselines~\cite{deng2025emerging, wu2025omnigen2, wu2025qwen}. Specifically, for OmniGen2~\cite{wu2025omnigen2}, we expand the image embeddings from 5 to 10 using normal initialization to accommodate a maximum of 10 input images.

\subsection{Detailed Quantitative Results on \benchname}
\input{secs/appendix/y-res}

We present detailed results on our \benchname in \cref{tab:appendix_customization,tab:appendix_illustration,tab:appendix_spatial,tab:appendix_temporal}, reporting the scores across different image count categories for each task. As demonstrated, open-source models exhibit a pronounced performance degradation when conditioned on more than three reference images. Notably, Qwen-Image-Edit-2511~\cite{wu2025qwen} suffers a catastrophic drop from 8.27 (1--3 references) to 4.55 (4--5 references) in the Customization task.

Furthermore, these detailed tables reveal that current open-source models and existing datasets~\cite{ye2025echo,wei2025mico,liu2025opensubject} struggle to support long-context multi-reference generation, even on their specifically targeted Customization tasks (\cref{tab:appendix_customization}). For instance, although MICo~\cite{wei2025mico} advocates for multi-reference generation, its training data---constructed via simple decomposition and recomposition---fails to provide the robust inter-reference reasoning capabilities necessary for models to process a large number of input images effectively.

Finally, the detailed metrics highlight a distinct contrast between progressive and non-progressive tasks. As shown, metrics for the progressive Customization task exhibit a clear descending trend as the number of input images increases. Conversely, performance on non-progressive tasks (Illustration, Spatial, and Temporal) remains comparatively stable across all image count categories for most baseline models. This pronounced degradation at higher image counts further underscores the necessity to increase the volume of 6--10 image samples, as implemented during our data construction.

\subsection{More Qualitative Results}
\input{secs/appendix/v-selection}

\subsubsection{\benchname Results of Different Tasks}

We display the generated results for each task in \benchname, produced by Bagel~\cite{deng2025emerging} fine-tuned on our \dataname, as illustrated in~\cref{fig:app_vis_cus,fig:app_vis_ill,fig:app_vis_spa,fig:app_vis_tmp}. In the \textit{Customization} task, the model demonstrates outstanding performance in composing multiple reference images into a single coherent and natural scene, as shown in~\cref{fig:app_vis_cus}, even when presented with more than 8 input images. In the \textit{Illustration} task, the model further exhibits the capability to generate complementary images that enrich and enhance interleaved textual descriptions. For the \textit{Spatial} task, the model demonstrates a strong understanding of 3D spatial relationships, successfully synthesizing novel views from specified viewpoints. Finally, in the \textit{Temporal} task, the model effectively captures the transformation patterns across video frames and generates plausible future scenes.

\subsubsection{Token Selection}

To further illustrate the differences among the token selection strategies, we present quantitative results for block-wise, image-wise, and text-aligned selection in~\cref{fig:selection_vis}. Specifically, the results employ a retention rate of 90\% for block-wise, 50\% for image-wise, and 30\% for text-aligned selection.

As shown, block-wise selection preserves high-quality generation; however, due to the dynamic dropping of certain tokens, the generated images occasionally appear unnatural, as evidenced by the fourth case. Image-wise selection primarily suffers from the degradation of cross-reference blending. Constrained by the retention rate, the generation tends to focus on specific images, particularly when the number of input images is small, as demonstrated in the first, second, and third cases, ultimately leading to failures in coherent blending. For instance, the woman's face disappears in the first case, the identity of the second individual is lost in the second case, and the jacket is not generated appropriately in the third case.

In contrast, despite retaining only 30\% of the tokens, text-aligned selection effectively preserves the most critical information, achieving competitive results across all cases. Nevertheless, due to the selective discarding of certain tokens, some fine-grained details are missing, including the flowers in the fifth case and the paintings in the sixth case.

%% file: secs/appendix/y-res.tex
\newcolumntype{Z}{>{\centering\arraybackslash}p{1.1cm}}

\begin{table*}[tbp]
\centering
\caption{\textbf{Detailed Results on Customization Task.}
  Columns 1--3, 4--5, 6--7, 8--10 denote difficulty bins;
  Avg is the macro-average over all bins.}
\vspace{-6pt}
\resizebox{0.95\linewidth}{!}{%
\begin{tabular}{l|ZZZZZ}
  \toprule
  \textbf{Model} & \textbf{1--3} & \textbf{4--5} & \textbf{6--7} & \textbf{8--10} & \textbf{Avg$\uparrow$} \\
  \midrule
  \rowcolor[HTML]{EFEFEF}\multicolumn{6}{c}{\textit{Closed-source models}} \\
  \midrule
  Nano Banana Pro~\cite{google2025nanobanana}    & 9.58 & 9.24 & 8.18 & 7.01 & 8.50 \\
  GPT-Image-1.5~\cite{openai2025gptimage1dot5}   & 9.67 & 9.57 & 9.30 & 8.23 & 9.19 \\
  \midrule
  \rowcolor[HTML]{EFEFEF}\multicolumn{6}{c}{\textit{Open-source models}} \\
  \midrule
  BAGEL~\cite{deng2025emerging}                             & 6.78 & 3.96 & 2.87 & 2.19 & 3.95 \\
  BAGEL + Echo4o~\cite{ye2025echo}                          & 7.89 & 6.01 & 4.71 & 3.24 & 5.46 \\
  BAGEL + MICo~\cite{wei2025mico}                           & 7.67 & 4.97 & 3.29 & 2.39 & 4.58 \\
  OmniGen2~\cite{wu2025omnigen2}                            & 6.66 & 3.76 & 2.55 & 1.99 & 3.74 \\
  OmniGen2 + MICo~\cite{wei2025mico}                        & 6.19 & 3.74 & 2.40 & 1.95 & 3.57 \\
  OmniGen2 + OpenSubject~\cite{liu2025opensubject}          & 6.47 & 3.70 & 2.54 & 2.13 & 3.71 \\
  Qwen-Image-Edit-2511~\cite{wu2025qwen}                    & 8.27 & 4.55 & 1.15 & 0.69 & 3.67 \\
  \midrule
  \rowcolor{cyan!5} Bagel + Ours    & \textbf{9.00} & \textbf{8.84} & \textbf{8.01} & \textbf{6.23} & \textbf{8.02} \\
  \rowcolor{cyan!5} OmniGen2 + Ours & 7.71 & 6.88 & 5.17 & 3.73 & 5.87 \\
  \rowcolor{cyan!5} Qwen + Ours     & 8.77 & 7.85 & 5.89 & 3.49 & 6.50 \\
  \bottomrule
\end{tabular}
}
\label{tab:appendix_customization}
\end{table*}

\begin{figure}[tbp]
    \vspace{-10pt}
    \centering
    \includegraphics[width=0.95\textwidth]{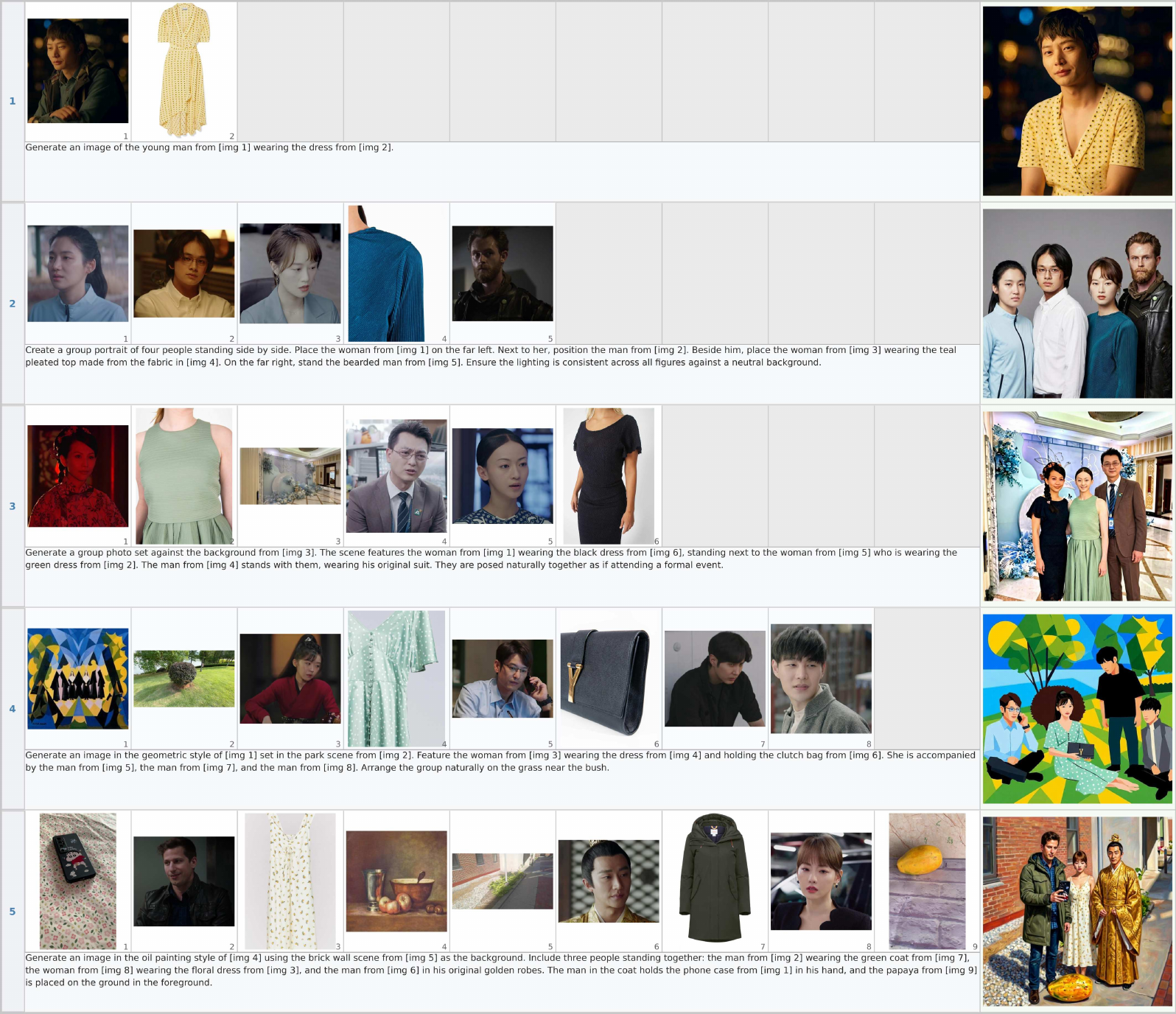}
    \caption{\textbf{Qualitative Results} of \benchname Customization tasks for Bagel~\cite{deng2025emerging} fine-tuned on \dataname.}
    \label{fig:app_vis_cus}
    \vspace{-10pt}
\end{figure}

\begin{table*}[tbp]
\centering
\caption{\textbf{Detailed Results on Illustration Task.}
  Columns 1--3, 4--5, 6--7, 8--10 denote difficulty bins;
  Avg is the macro-average over all bins.}
\vspace{-6pt}
\resizebox{0.95\linewidth}{!}{%
\begin{tabular}{l|ZZZZZ}
  \toprule
  \textbf{Model} & \textbf{1--3} & \textbf{4--5} & \textbf{6--7} & \textbf{8--10} & \textbf{Avg$\uparrow$} \\
  \midrule
  \rowcolor[HTML]{EFEFEF}\multicolumn{6}{c}{\textit{Closed-source models}} \\
  \midrule
  Nano Banana Pro~\cite{google2025nanobanana}    & 9.39 & 8.85 & 8.87 & 8.89 & 9.00 \\
  GPT-Image-1.5~\cite{openai2025gptimage1dot5}   & 9.41 & 8.39 & 8.77 & 8.91 & 8.87 \\
  \midrule
  \rowcolor[HTML]{EFEFEF}\multicolumn{6}{c}{\textit{Open-source models}} \\
  \midrule
  BAGEL~\cite{deng2025emerging}                             & 4.66 & 4.35 & 4.54 & 4.14 & 4.42 \\
  BAGEL + Echo4o~\cite{ye2025echo}                          & 4.71 & 4.32 & 4.14 & 4.13 & 4.33 \\
  BAGEL + MICo~\cite{wei2025mico}                           & 4.81 & 4.77 & 4.68 & 4.56 & 4.70 \\
  OmniGen2~\cite{wu2025omnigen2}                            & 5.08 & 4.36 & 4.02 & 3.58 & 4.26 \\
  OmniGen2 + MICo~\cite{wei2025mico}                        & 4.78 & 4.21 & 4.20 & 3.76 & 4.24 \\
  OmniGen2 + OpenSubject~\cite{liu2025opensubject}          & 4.90 & 4.06 & 3.77 & 3.50 & 4.06 \\
  Qwen-Image-Edit-2511~\cite{wu2025qwen}                    & \textbf{5.79} & 3.58 & 2.68 & 1.65 & 3.43 \\
  \midrule
  \rowcolor{cyan!5} Bagel + Ours    & 5.78 & \textbf{5.59} & \textbf{5.50} & \textbf{5.62} & \textbf{5.62} \\
  \rowcolor{cyan!5} OmniGen2 + Ours & 4.61 & 4.74 & 4.18 & 4.31 & 4.46 \\
  \rowcolor{cyan!5} Qwen + Ours     & 5.46 & 4.62 & 3.93 & 3.04 & 4.26 \\
  \bottomrule
\end{tabular}
}
\label{tab:appendix_illustration}
\end{table*}

\begin{figure}[tbp]
    \vspace{-10pt}
    \centering
    \includegraphics[width=0.95\textwidth]{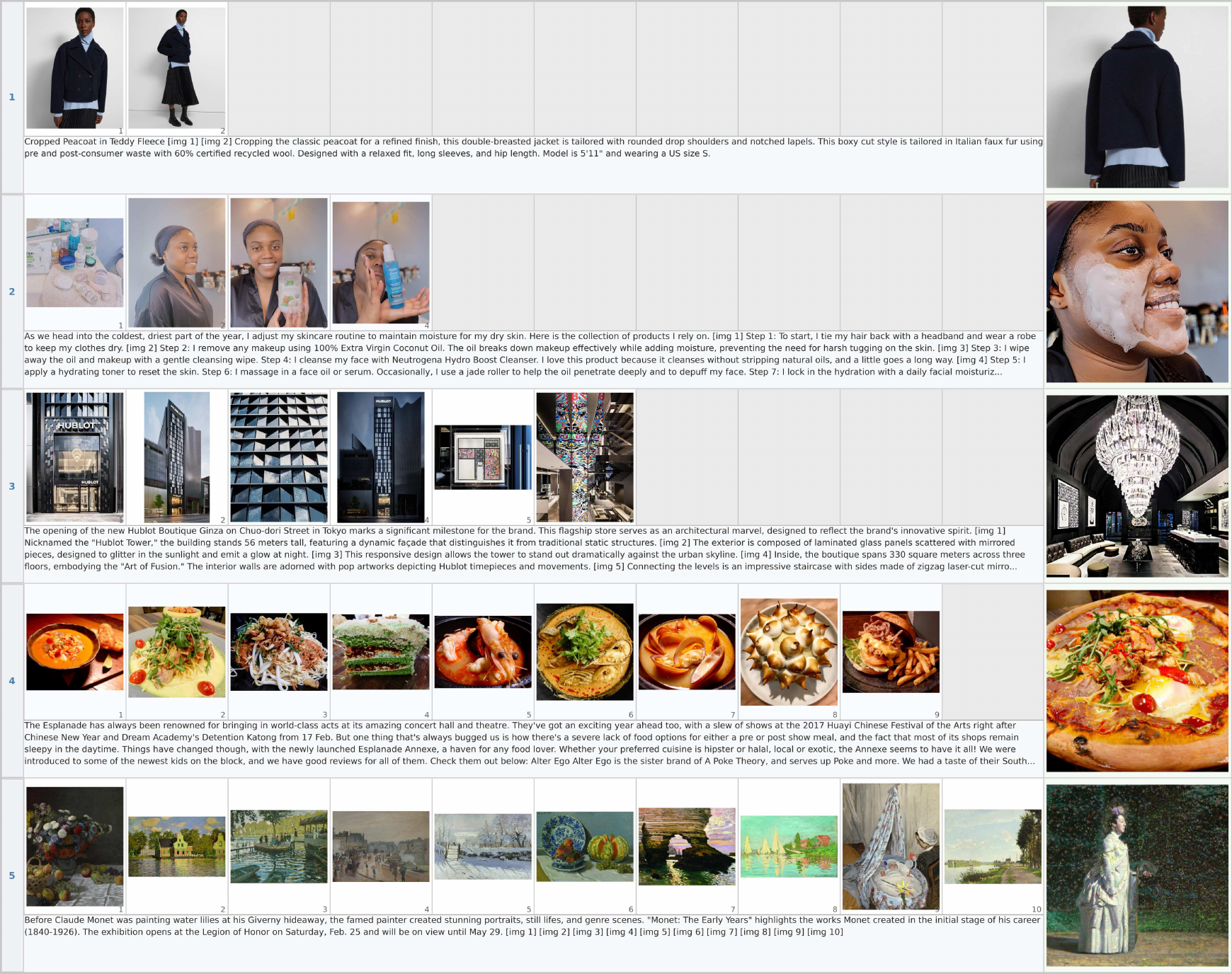}
    \caption{\textbf{Qualitative Results} of \benchname Illustration tasks for Bagel~\cite{deng2025emerging} fine-tuned on \dataname.}
    \label{fig:app_vis_ill}
    \vspace{-10pt}
\end{figure}

\begin{table*}[tbp]
\centering
\caption{\textbf{Detailed Results on Spatial Task.}
  Columns 1--3, 4--5, 6--7, 8--10 denote difficulty bins;
  Avg is the macro-average over all bins.}
\vspace{-6pt}
\resizebox{0.95\linewidth}{!}{%
\begin{tabular}{l|ZZZZZ}
  \toprule
  \textbf{Model} & \textbf{1--3} & \textbf{4--5} & \textbf{6--7} & \textbf{8--10} & \textbf{Avg$\uparrow$} \\
  \midrule
  \rowcolor[HTML]{EFEFEF}\multicolumn{6}{c}{\textit{Closed-source models}} \\
  \midrule
  Nano Banana Pro~\cite{google2025nanobanana}    & 3.09 & 3.47 & 3.34 & 3.05 & 3.24 \\
  GPT-Image-1.5~\cite{openai2025gptimage1dot5}   & 3.01 & 3.63 & 4.25 & 4.23 & 3.78 \\
  \midrule
  \rowcolor[HTML]{EFEFEF}\multicolumn{6}{c}{\textit{Open-source models}} \\
  \midrule
  BAGEL~\cite{deng2025emerging}                             & 0.87 & 0.47 & 0.53 & 0.53 & 0.60 \\
  BAGEL + Echo4o~\cite{ye2025echo}                          & 0.88 & 0.63 & 0.90 & 0.70 & 0.78 \\
  BAGEL + MICo~\cite{wei2025mico}                           & 0.97 & 0.64 & 0.87 & 0.73 & 0.80 \\
  OmniGen2~\cite{wu2025omnigen2}                            & 0.61 & 0.72 & 0.94 & 1.08 & 0.84 \\
  OmniGen2 + MICo~\cite{wei2025mico}                        & 0.51 & 0.55 & 0.93 & 1.01 & 0.75 \\
  OmniGen2 + OpenSubject~\cite{liu2025opensubject}          & 1.07 & 1.04 & 1.52 & 1.50 & 1.28 \\
  Qwen-Image-Edit-2511~\cite{wu2025qwen}                    & 1.91 & 1.23 & 0.73 & 0.52 & 1.10 \\
  \midrule
  \rowcolor{cyan!5} Bagel + Ours    & \textbf{3.40} & \textbf{3.24} & \textbf{3.21} & \textbf{3.74} & \textbf{3.40} \\
  \rowcolor{cyan!5} OmniGen2 + Ours & 1.65 & 1.98 & 1.32 & 1.47 & 1.60 \\
  \rowcolor{cyan!5} Qwen + Ours     & 2.82 & 2.69 & 2.57 & 2.63 & 2.68 \\
  \bottomrule
\end{tabular}
}
\label{tab:appendix_spatial}
\end{table*}

\begin{figure}[tbp]
    \vspace{-10pt}
    \centering
    \includegraphics[width=0.95\textwidth]{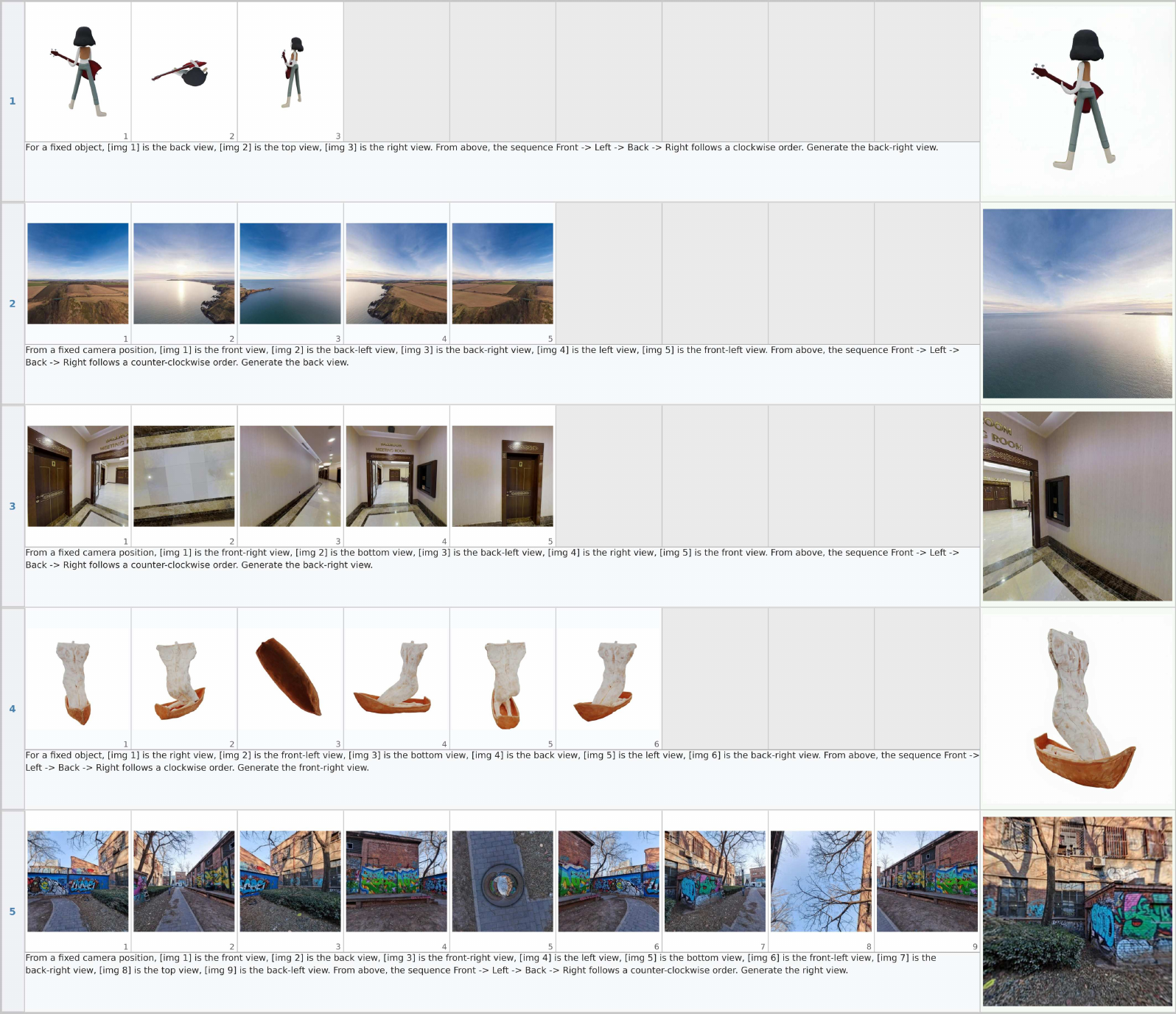}
    \caption{\textbf{Qualitative Results} of \benchname Spatial tasks for Bagel~\cite{deng2025emerging} fine-tuned on \dataname.}
    \label{fig:app_vis_spa}
    \vspace{-10pt}
\end{figure}

\begin{table*}[tbp]
\centering
\caption{\textbf{Detailed Results on Temporal Task.}
  Columns 1--3, 4--5, 6--7, 8--10 denote difficulty bins;
  Avg is the macro-average over all bins.}
\vspace{-6pt}
\resizebox{0.95\linewidth}{!}{%
\begin{tabular}{l|ZZZZZ}
  \toprule
  \textbf{Model} & \textbf{1--3} & \textbf{4--5} & \textbf{6--7} & \textbf{8--10} & \textbf{Avg$\uparrow$} \\
  \midrule
  \rowcolor[HTML]{EFEFEF}\multicolumn{6}{c}{\textit{Closed-source models}} \\
  \midrule
  Nano Banana Pro~\cite{google2025nanobanana}    & 8.62 & 7.79 & 7.28 & 7.21 & 7.73 \\
  GPT-Image-1.5~\cite{openai2025gptimage1dot5}   & 8.73 & 8.23 & 8.06 & 7.62 & 8.16 \\
  \midrule
  \rowcolor[HTML]{EFEFEF}\multicolumn{6}{c}{\textit{Open-source models}} \\
  \midrule
  BAGEL~\cite{deng2025emerging}                             & 3.90 & 2.82 & 3.13 & 2.72 & 3.14 \\
  BAGEL + Echo4o~\cite{ye2025echo}                          & 3.71 & 2.67 & 2.60 & 2.21 & 2.80 \\
  BAGEL + MICo~\cite{wei2025mico}                           & 4.30 & 3.40 & 3.70 & 3.36 & 3.69 \\
  OmniGen2~\cite{wu2025omnigen2}                            & 3.52 & 2.48 & 2.56 & 2.26 & 2.71 \\
  OmniGen2 + MICo~\cite{wei2025mico}                        & 3.13 & 2.49 & 2.61 & 2.28 & 2.63 \\
  OmniGen2 + OpenSubject~\cite{liu2025opensubject}          & 3.63 & 2.79 & 2.86 & 2.38 & 2.92 \\
  Qwen-Image-Edit-2511~\cite{wu2025qwen}                    & 4.54 & 3.49 & 1.43 & 0.84 & 2.58 \\
  \midrule
  \rowcolor{cyan!5} Bagel + Ours    & \textbf{6.34} & \textbf{6.12} & \textbf{5.64} & \textbf{5.15} & \textbf{5.81} \\
  \rowcolor{cyan!5} OmniGen2 + Ours & 4.95 & 4.98 & 4.16 & 3.97 & 4.51 \\
  \rowcolor{cyan!5} Qwen + Ours     & 5.66 & 4.40 & 3.60 & 2.35 & 4.00 \\
  \bottomrule
\end{tabular}
}
\label{tab:appendix_temporal}
\end{table*}

\begin{figure}[tbp]
    \vspace{-10pt}
    \centering
    \includegraphics[width=0.95\textwidth]{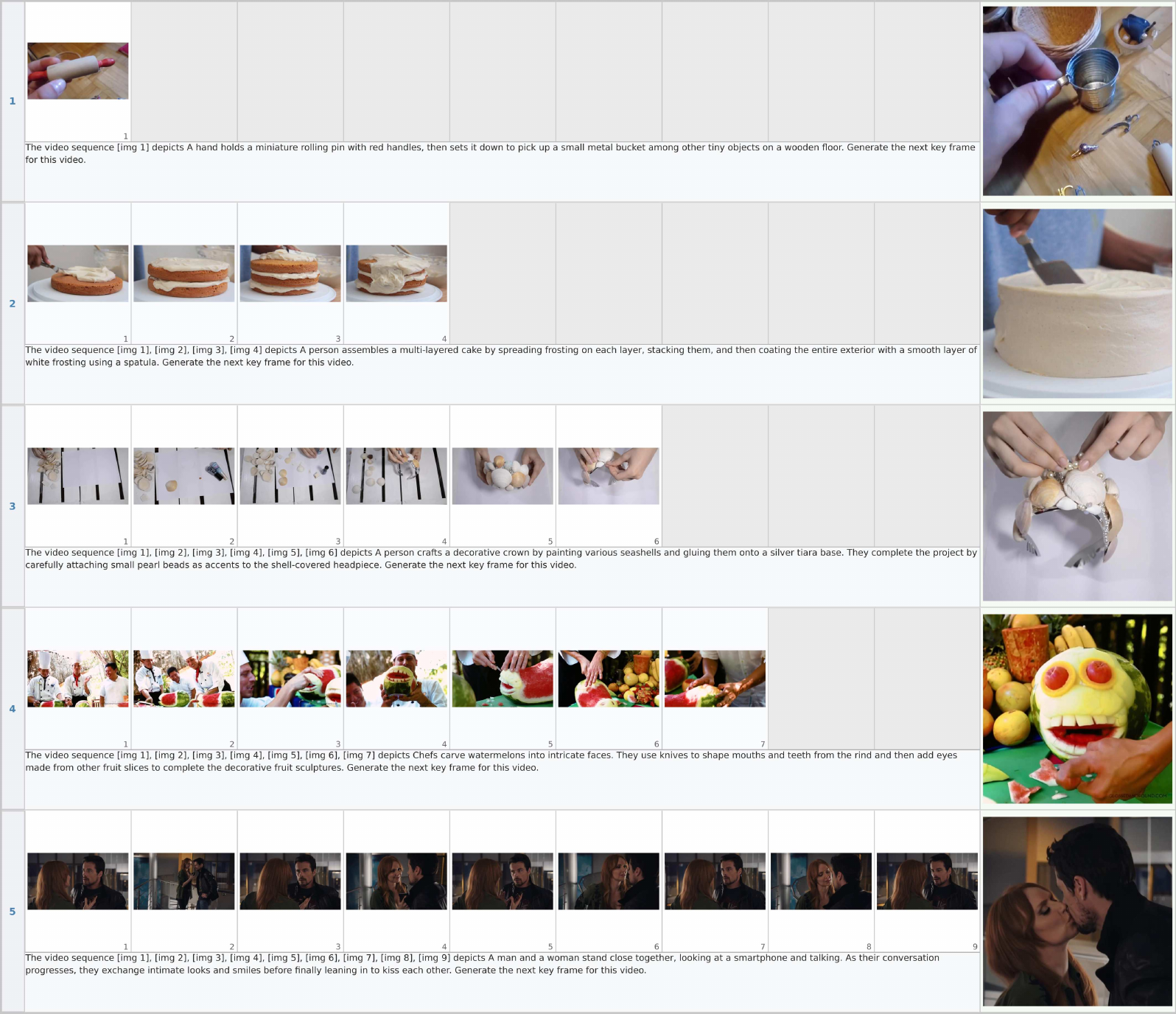}
    \caption{\textbf{Qualitative Results} of \benchname Temporal tasks for Bagel~\cite{deng2025emerging} fine-tuned on \dataname.}
    \label{fig:app_vis_tmp}
    \vspace{-10pt}
\end{figure}

%% file: secs/appendix/v-selection.tex
\begin{figure}[tbp]
    \vspace{-10pt}
    \centering
    \includegraphics[width=0.95\textwidth]{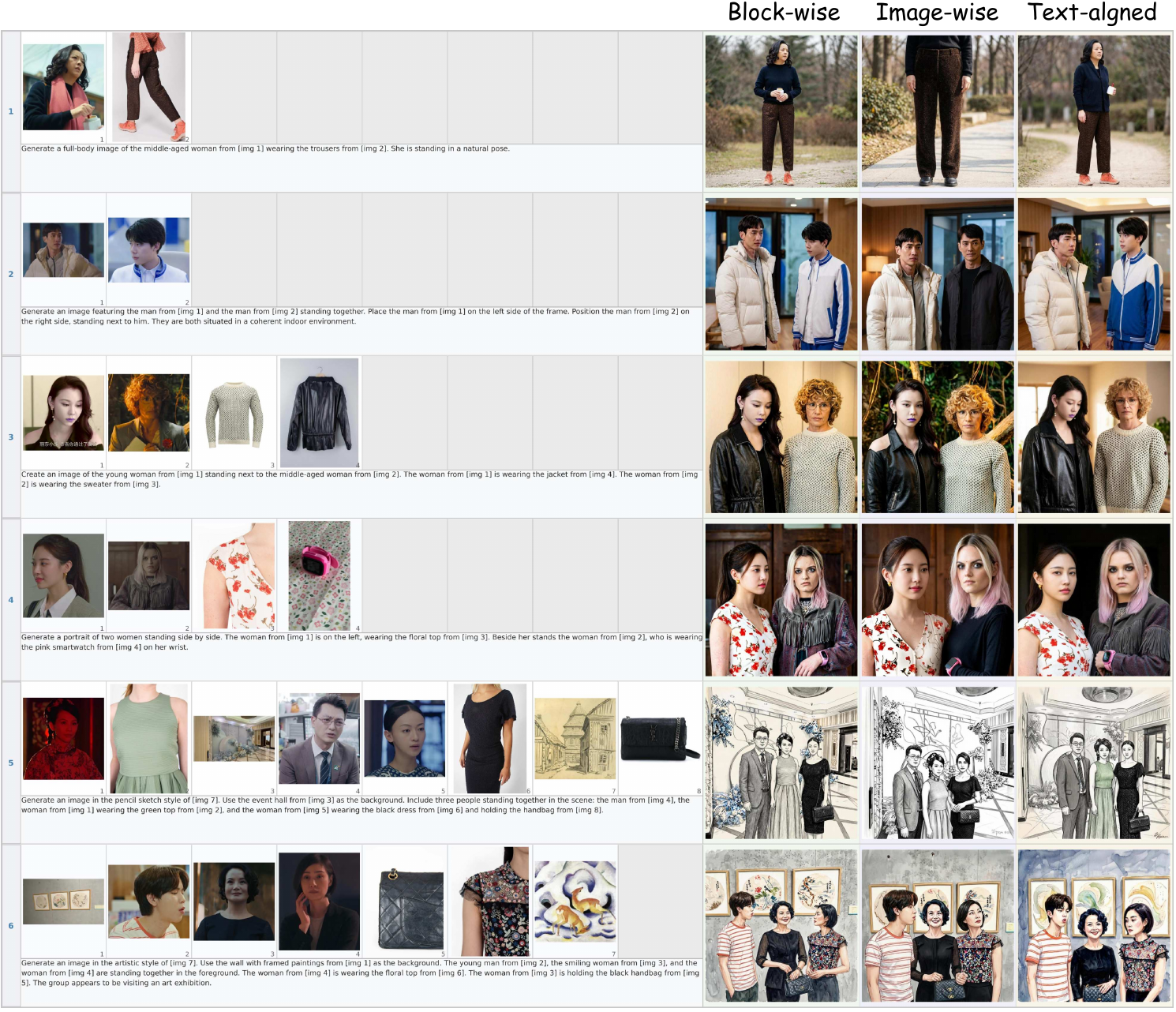}
    \caption{\textbf{Quantitative Results} of different token selection strategies.}
    \label{fig:selection_vis}
    \vspace{-10pt}
\end{figure}

%% file: secs/appendix/4-failure.tex
\section{Failure Cases}

\input{secs/appendix/z-failure}

We present failure cases of Bagel~\cite{deng2025emerging} fine-tuned on our \dataname in~\cref{fig:failure_case}, including \textcolor{blue}{blue} input images as references, instructions, \textcolor{green}{green} ground truths (no GT is provided for customization), and \textcolor{red}{red} generated failure results.

\subsubsection{Customization} primarily suffers from the reference number problem. As shown in~\cref{fig:failure_case}(a), when too many input images are provided, the model tends to ``forget'' certain references. For instance, the fourth woman in this case disappears, and an output with only three humans instead of the required four is generated. This consequently leads to the disappearance of clothing associated with the omitted identity, and introduces identity mixing to some extent, like the hair bundle of the fourth woman is incorrectly blended onto the second woman.

\subsubsection{Illustration} exhibits problems related to long-context information gathering and text rendering. As displayed in~\cref{fig:failure_case}(b), the original context requires generating accessories that appeared in the first image. However, the model fails to retrieve such information and instead attempts to generate new, incorrect ones. Furthermore, as the model has not been specifically trained for text rendering, the generated text is garbled and semantically meaningless.

\subsubsection{Spatial} requires the model to understand the relationships among different viewpoints. However, the model sometimes struggles with spatial relationships. In~\cref{fig:failure_case}(c), the first image depicts a back view, and the target requires a back-left view, meaning the viewpoint should shift to the right relative to the first image. However, the model misinterprets this and shifts to the left, resulting in an incorrect generation direction.

\subsubsection{Temporal} asks the model to maintain frame-sequence consistency and predict future states. However, handling long preceding contexts remains challenging. For example, in~\cref{fig:failure_case}(d), the model incorrectly generates the clothing of the closer player as black, despite the player wearing white. Additionally, the scene is visually distorted, and the scoreboard is missing. These errors likely stem from the model's difficulty in capturing fine-grained details within long contexts, such as small visual concepts like the scoreboard and clothing color.

These failure cases reveal several deficiencies in the model's capabilities, including context retention, contextual consistency, text rendering, 3D spatial reasoning, and fine-detail perception. We regard these as important directions for future improvement.

%% file: secs/appendix/z-failure.tex
\begin{figure}[tbp]
    \vspace{-10pt}
    \centering
    \includegraphics[width=0.95\textwidth]{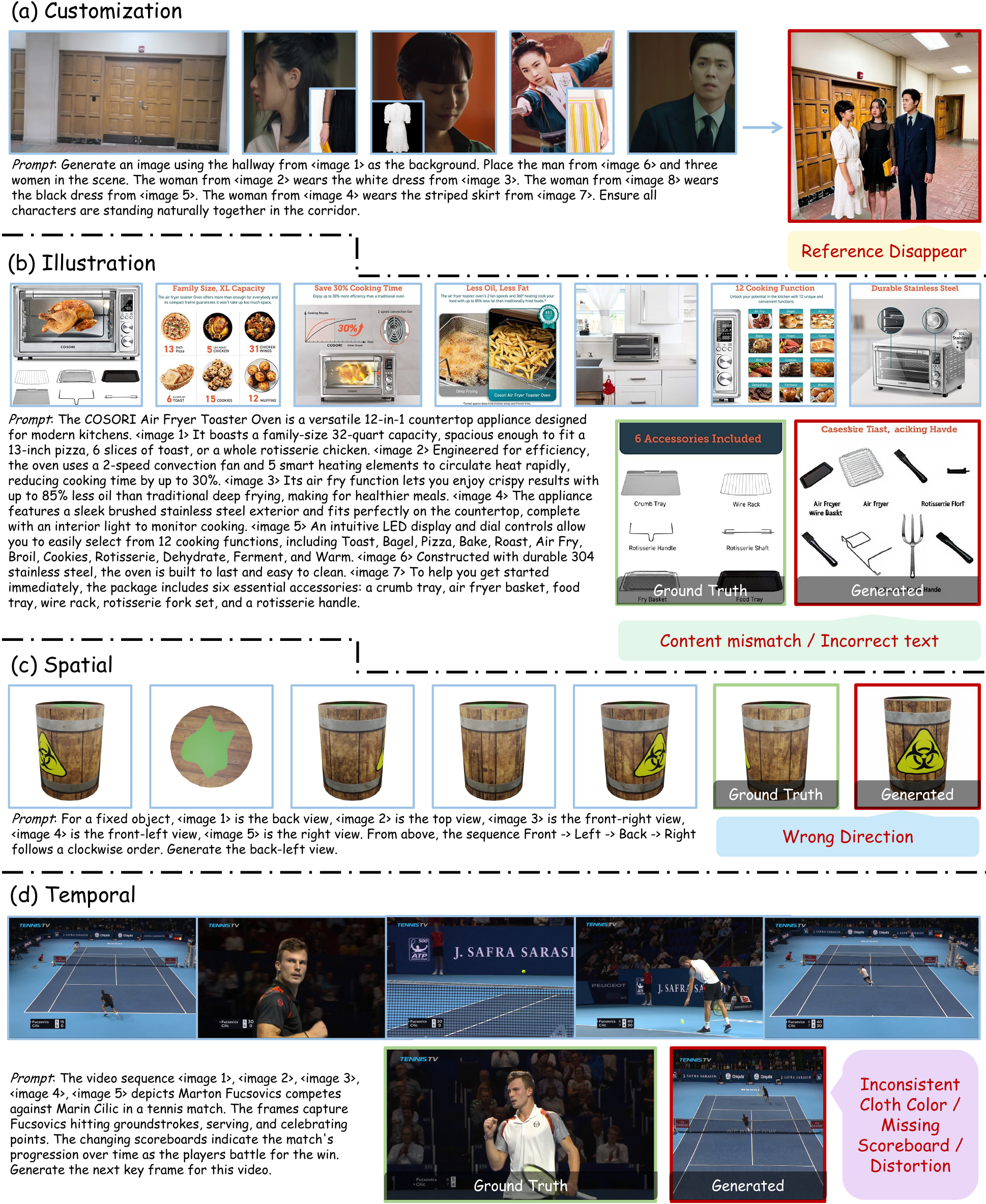}
    \caption{\textbf{Failure Cases} of Bagel~\cite{deng2025emerging} fine-tuned on our \dataname. Blue images are input references, green ones are ground truth, and red ones are generated failure results.}
    \label{fig:failure_case}
    \vspace{-20pt}
\end{figure}

%% file: secs/appendix/5-others.tex
\section{Limitation}
While \dataname significantly enhances multi-reference generation, our approach still exhibits performance degradation when scaling up to 6-10 input images, indicating that processing highly complex, long-context visual dependencies remains a challenge. Furthermore, our proposed \benchname, while a crucial first step, is still preliminary and covers a relatively limited range of predefined tasks; a more comprehensive and general evaluation framework is needed to fully assess generation capabilities in the wild. Finally, there remains a noticeable performance gap between our fine-tuned models and SOTA closed-source models, highlighting the need for further exploration in data scaling and model architectures.

\section{Social Impact}
The advancement of long-context multi-reference image generation holds significant potential to benefit creative domains requiring complex visual composition. However, it concurrently introduces dual-use risks, such as the generation of deceptive content or unauthorized identity manipulation. To mitigate potential ethical and legal concerns at the foundational level, our dataset construction strictly relies on publicly available sources and adheres to standard permissible licenses, aiming to minimize privacy risks and intellectual property disputes. Furthermore, to prevent negative social impacts during future model deployment, we advocate for the integration of robust technical safeguards, such as real-time output assessment and automated privacy detection mechanisms, to proactively identify and restrict malicious misuse during the generation process.

\section{Future Work}
Future research will focus on expanding \dataname to encompass a broader and more general range of multi-image scenarios. By scaling up the dataset to include samples with even more reference images, we aim to further increase the upper limit of input capacity and ultimately bridge the performance gap with state-of-the-art closed-source models. Concurrently, we plan to refine \benchname into a more granular evaluation framework, exploring the integration of detailed scoring methodologies, such as checklist-based assessments, to capture more nuanced generative alignments.

Additionally, a critical direction involves exploring advanced methodologies that enable models to more effectively utilize dense multi-image information. This includes building upon our preliminary explorations of token selection to design specialized token representations and attention mechanisms that are natively optimized for in-context generation, thereby maximizing both computational efficiency and generation performance.

%% file: secs/appendix/w-data.tex
\begin{figure}[t]
    \vspace{-10pt}
    \centering
    \includegraphics[width=0.95\textwidth]{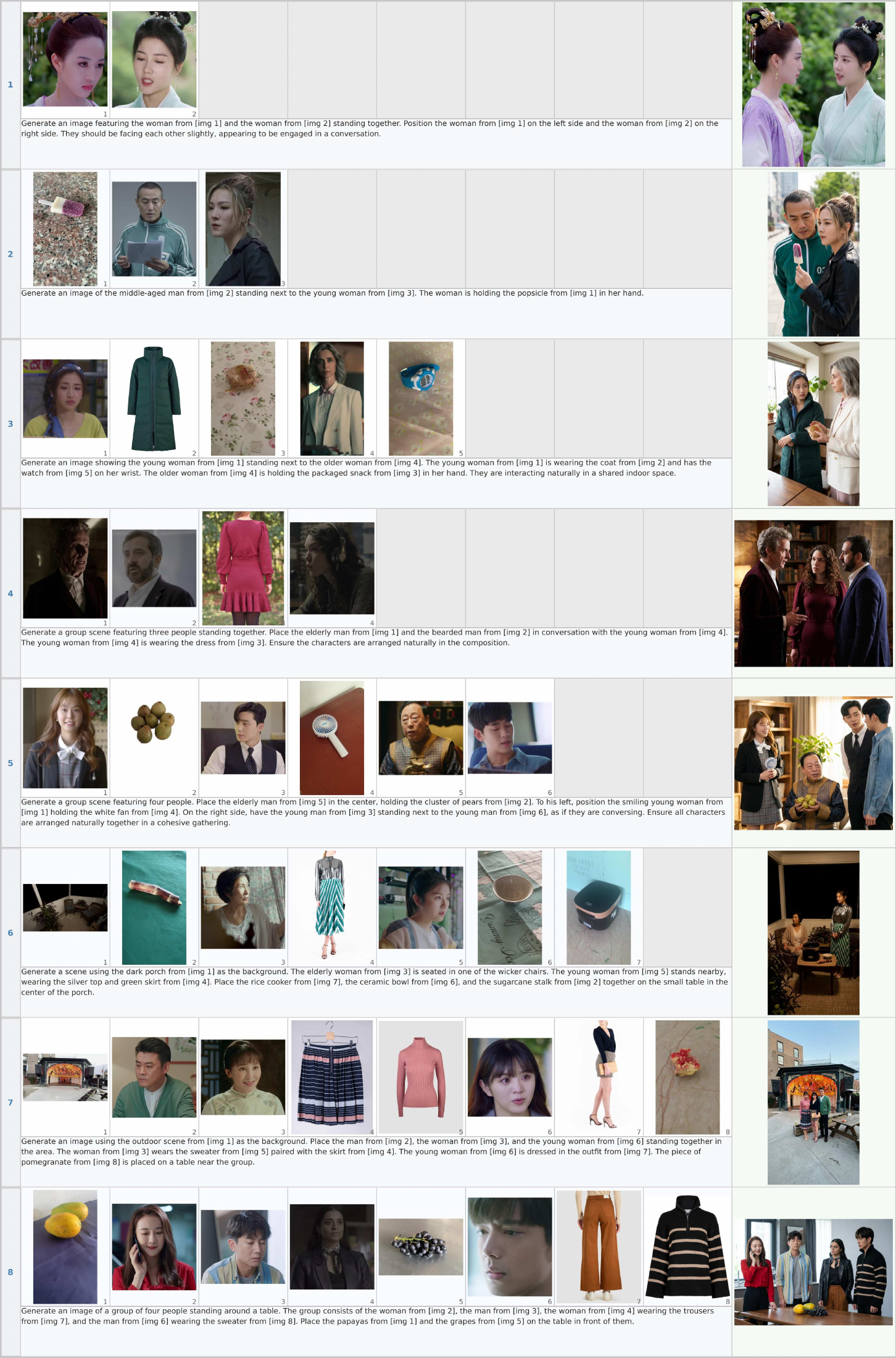}
    \caption{\textbf{Visualization} of \dataname Customization subset.}
    \label{fig:app_data_cus}
    \vspace{-20pt}
\end{figure}

\begin{figure}[t]
    \vspace{-10pt}
    \centering
    \includegraphics[width=0.95\textwidth]{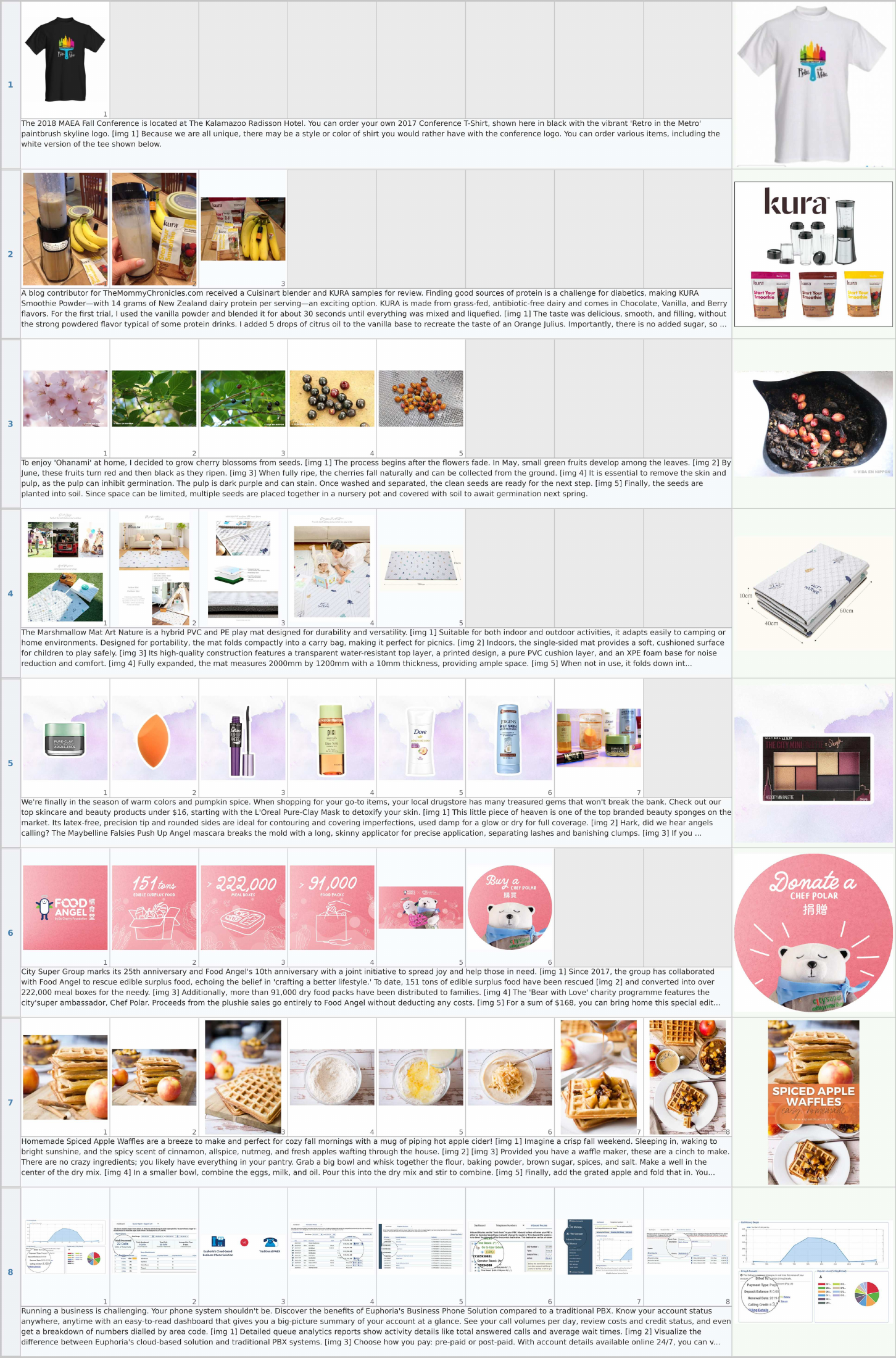}
    \caption{\textbf{Visualization} of \dataname Illustration subset.}
    \label{fig:app_data_ill}
    \vspace{-20pt}
\end{figure}

\begin{figure}[t]
    \vspace{-10pt}
    \centering
    \includegraphics[width=0.95\textwidth]{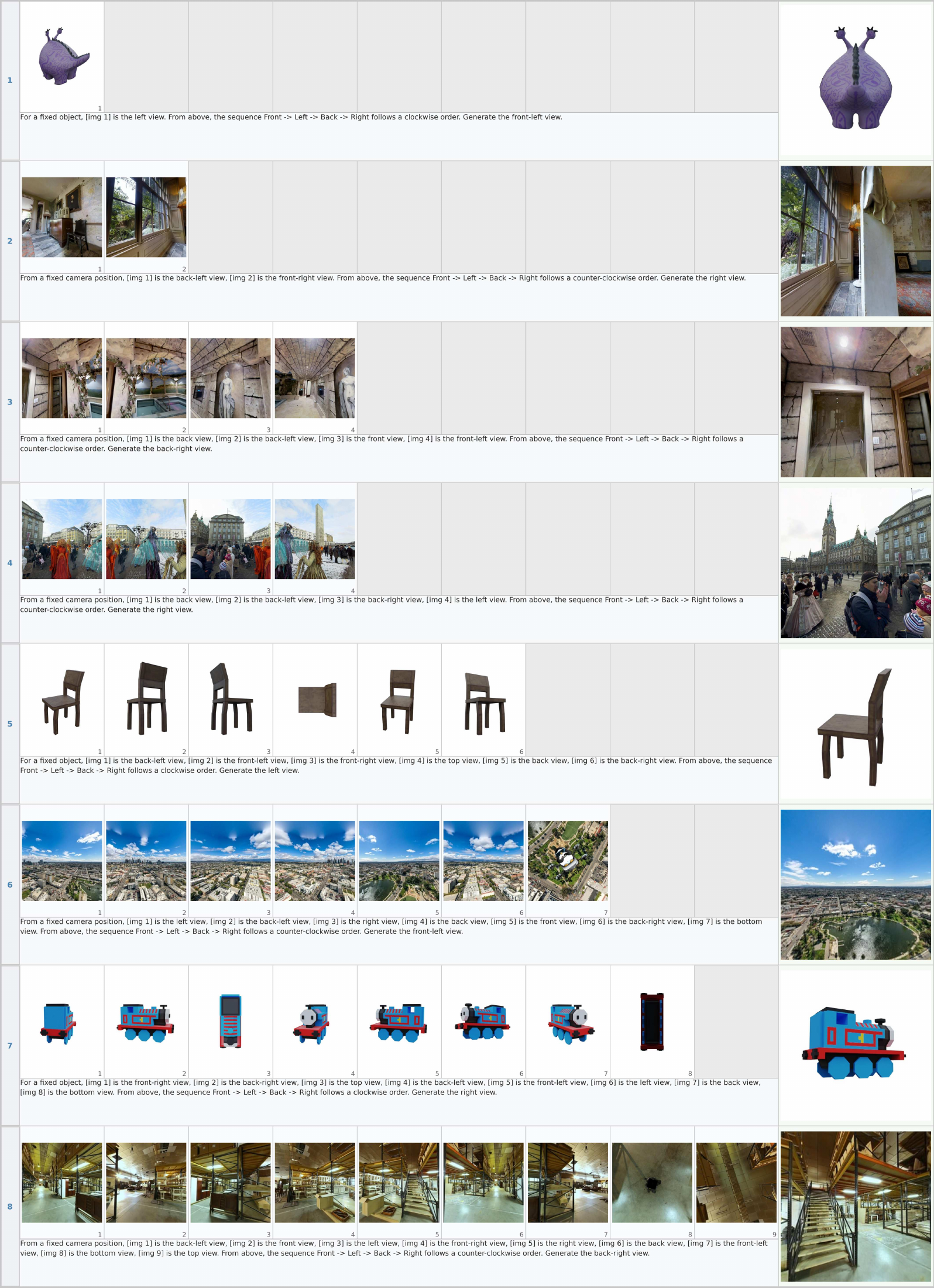}
    \caption{\textbf{Visualization} of \dataname Spatial subset.}
    \label{fig:app_data_spa}
    \vspace{-20pt}
\end{figure}

\begin{figure}[t]
    \vspace{-10pt}
    \centering
    \includegraphics[width=0.95\textwidth]{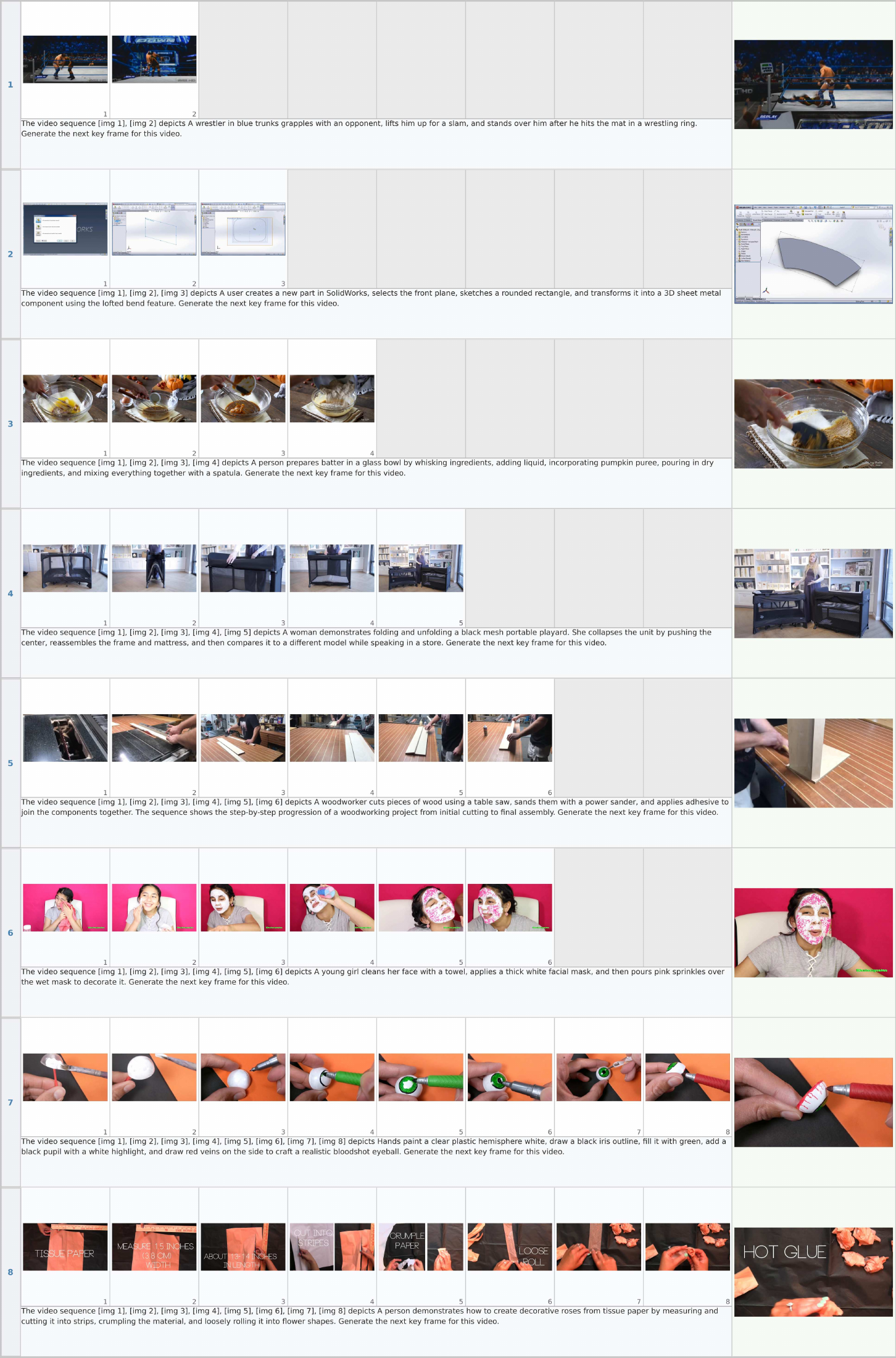}
    \caption{\textbf{Visualization} of \dataname Temporal subset.}
    \label{fig:app_data_tmp}
    \vspace{-20pt}
\end{figure}

%% file: secs/appendix/x-prompt.tex
\begin{figure*}[t!]
    \centering

    \begin{tcolorbox}[width=\textwidth, title={\textbf{Evaluation Prompt for Customization}}]
    \small

    You are a meticulous Digital Art Critic and Quality Assurance Specialist. You are known for your extremely high standards and attention to microscopic details. \\
    
    \textbf{1. THE INSTRUCTION (Read this first)} \\
    The AI model was given this specific customization instruction: \texttt{\{instruction\}} \\
    \textbf{2. THE IMAGES} \\
    You will be provided with a sequence of images. \\
    \textbf{Reference Images:} The first few images showing the subject/object to be customized. \\
    \textbf{Generated Image:} The \textbf{LAST} image in the sequence. This is the output you must evaluate. \\
    \texttt{\{image\_descriptions\}} \\
    \textbf{3. EVALUATION TASK} \\
    Evaluate the \textbf{Generated Image} (the last image) against the \textbf{Reference Images} and the \textbf{Instruction}. Your response must be a JSON object: \\
    \texttt{\{"consistency\_scores": [<score1>, ..., <scoreN>], "following\_score": <score>, "overall\_reasoning": "..."\}} \\
    \textbf{4. CRITICAL CONFLICT RESOLUTION} \\
    \textbf{Style/Attribute Changes:} If the instruction explicitly asks to change a feature (e.g., ``make him old'', ``change hair color''), \textbf{DO NOT penalize} the Consistency Score for these requested changes. \\
    \textbf{Unintended Changes:} Penalize strictly if features change \textit{without} being asked. \\
    \textbf{5. SCORING CRITERIA} \\
    \textit{Metric 1 --- Image Consistency Score (ICS), 0--10, one per reference image:}
    Does the generated image preserve the identity of subjects/objects from each reference image?\\
    \textbf{10:} Identity perfectly preserved.
    \textbf{8--9:} Clearly recognizable; minor details differ slightly.
    \textbf{5--7:} Resembles a ``look-alike'' rather than the exact subject; key ratios slightly off.
    \textbf{2--4:} Vague resemblance; specific identity is lost.
    \textbf{0--1:} Completely different person or object. \\
    Focus on: faces (eye shape, nose structure, jawline), object markings/logos, and clothing unless the instruction implies a change. \\

    \textit{Metric 2 --- Prompt Following Score (PFS), 0--10:}
    Does the generated image fulfill the editing instruction accurately?\\
    \textbf{10:} All aspects executed perfectly; no hallucinations.
    \textbf{8--9:} Follows main instruction; misses a tiny detail.
    \textbf{5--7:} Captures general idea but gets details wrong or ignores secondary constraints.
    \textbf{2--4:} Misses the main point or hallucinates significant content.
    \textbf{0--1:} Completely ignores the instruction.\\
    \end{tcolorbox}

    \caption{Evaluation prompt for the Customization task.}
    \label{fig:prompt_customization}
\end{figure*}
\clearpage

\begin{figure*}[t!]
    \centering

    \begin{tcolorbox}[width=\textwidth, title={\textbf{Evaluation Prompt for Illustration}}]

    You are an expert Visual Communication Specialist and Content Quality Auditor capable of evaluating diverse visual content including artistic illustrations, diary entries, manuals, advertisements, and presentation slides. \\

    \textbf{1. THE TASK CONTEXT (Read this first)} \\
    This is a \textbf{Visual Content Generation Task}. The AI model was given a text description and reference images, and asked to generate an image suitable to be placed after the context. \\
    Text description: \texttt{\{instruction\}} \\

    \textbf{2. THE IMAGES} \\
    \texttt{\{image\_descriptions\}} \\
    Image layout: $[\underbrace{\text{Ref}_1, \ldots, \text{Ref}_N}_{\text{context sequence}}\ |\ \text{Generated}]$ \\

    \textbf{3. EVALUATION TASK} \\
    Evaluate the \textbf{Generated Image} against the \textbf{Text Description} and \textbf{Reference Images} (if any). Return: \\
    \texttt{\{"text\_consistency\_score": <score>, "image\_consistency\_score": <score>, "overall\_reasoning": "..."\}} \\

    \textbf{4. SCORING CRITERIA} \\
    \textit{Metric 1 --- Text Consistency Score (TCS), 0--10:}
    Does the generated image accurately reflect the content, format, and intent of the text?\\
    \textbf{10:} Perfectly matches content, format, and style; well integrated with the context.
    \textbf{8--9:} Follows context well with only minor deviations.
    \textbf{5--7:} Captures main topic but fails in format or context requirements.
    \textbf{2--4:} Fails to represent core content or ignores format instructions.
    \textbf{0--1:} Irrelevant to the text. \\

    \textit{Metric 2 --- Image Consistency Score (ICS), 0--10:}
    Does the generated image correctly utilize visual information from the reference images?
    Note: references convey subject identity, \textbf{not necessarily style}, unless explicitly requested.
    \textit{With references:}\\
    \textbf{10:} Subjects/objects from references correctly featured; key visual features preserved.
    \textbf{8--9:} Clearly recognizable; minor deviations in non-essential details.
    \textbf{5--7:} Somewhat resembles reference; significant features distorted or missing.
    \textbf{2--4:} Barely recognizable; looks like a generic version.
    \textbf{0--1:} Reference images completely ignored. \\
    \textit{Without references:}
    \textbf{10:} Perfect internal consistency; logical lighting and spatial relationships.
    \textbf{8--9:} Good internal consistency with minor logical flaws.
    \textbf{5--7:} Noticeable internal contradictions or broken visual logic.
    \textbf{0--4:} Incoherent or chaotic.

    \end{tcolorbox}

    \caption{Evaluation prompt for the Illustration task.}
    \label{fig:prompt_illustration}
\end{figure*}
\clearpage

\begin{figure*}[t!]
    \centering

    \begin{tcolorbox}[width=\textwidth, title={\textbf{Evaluation Prompt for Spatial}}]

    You are a meticulous 3D Quality Assurance Specialist and Digital Art Critic known for your extremely high standards and attention to microscopic detail. \\

    \textbf{1. THE INSTRUCTION (Read this first)} \\
    The AI model was given this specific spatial instruction: \texttt{\{instruction\}} \\

    \textbf{2. THE IMAGES} \\
    \texttt{\{image\_descriptions\}} \\
    Image layout: $[\underbrace{\text{Ref}_1, \ldots, \text{Ref}_N}_{\text{input viewpoints}}\ |\ \text{Target (GT)}\ |\ \text{Generated}]$ \\

    \textbf{3. EVALUATION TASK} \\
    Compare the \textbf{Generated Image} against the \textbf{Reference Images} and the \textbf{Target Image} (Ground Truth). Return: \\
    \texttt{\{"viewpoint\_transformation\_score": <score>, "content\_consistency\_score": <score>, "overall\_reasoning": "..."\}} \\

    \textbf{4. SCORING CRITERIA (STRICT MODE)} \\
    \textit{Metric 1 --- View Transformation Score (VTS), 0--10:}
    Did the model move the camera \textit{exactly} according to the instruction?\\
    \textbf{10:} Viewpoint matches the Target Image geometry exactly.
    \textbf{8--9:} Movement correct; very minor deviation in angle, elevation, or distance.
    \textbf{5--7:} Generally correct direction but wrong magnitude (e.g., 90° instead of 45°).
    \textbf{2--4:} Camera moved in the \textbf{wrong direction} (e.g., left instead of right) --- functional failure.
    \textbf{0--1:} Failed to transform the view (image identical to input) or image is broken. \textbf{Zero tolerance for lazy copying.} \\

    \textit{Metric 2 --- Content Consistency Score (CCS), 0--10:}
    Does the image preserve the object's identity, texture fidelity, and structural logic?
    \textbf{Critical logic:} (1) Visible regions must be preserved perfectly --- blur, texture loss, or color shift incur severe penalty. (2) Invisible/occluded regions must be geometrically precise and stylistically seamless. \\
    \textbf{10:} Indistinguishable from Ground Truth; sharp textures, perfect geometry, no artifacts.
    \textbf{8--9:} Preserves identity well; minor loss of high-frequency texture or slight lighting inconsistency.
    \textbf{5--7:} Visible parts preserved but softer/blurrier than reference; invisible regions differ from target but are plausible.
    \textbf{2--4:} Significant structural warping; textures washed out; object identity barely maintained.
    \textbf{0--1:} Hallucination of a completely different object or severe visual noise. \\

    Explicitly mention what defects caused the score reduction (e.g., ``Texture on the armrest is blurry'', ``Camera angle is 15 degrees too shallow'').

    \end{tcolorbox}

    \caption{Evaluation prompt for the Spatial task.}
    \label{fig:prompt_spatial}
\end{figure*}
\clearpage

\begin{figure*}[t!]
    \centering

    \begin{tcolorbox}[width=\textwidth, title={\textbf{Evaluation Prompt for Temporal}}]

    You are a strict Video Continuity Specialist and QA Expert. Your task is to evaluate the quality of a generated frame in a temporal sequence. \\

    \textbf{1. THE INSTRUCTION} \\
    The specific action or change required for this frame: \texttt{\{instruction\}} \\
    \textbf{2. THE IMAGES} \\
    \texttt{\{image\_descriptions\}} \\
    Image layout: $[\underbrace{\text{Ref}_1, \ldots, \text{Ref}_N}_{\text{previous frames}}\ |\ \text{Generated (next frame)}]$ \\
    \textbf{3. EVALUATION TASK} \\
    Compare the \textbf{Generated Image} against the \textbf{Reference Images} (especially the last frame) and the \textbf{Instruction}. Return: \\
    \texttt{\{"context\_consistency\_score": <score>, \\"image\_sequence\_consistency\_score": <score>, \\"overall\_reasoning": "..."\}} \\
    
    \textbf{4. SCORING CRITERIA} \\
    \textit{Metric 1 --- Context \& Action Logic Score (CCS), 0--10:}
    Does the image correctly execute the text instruction and logically follow the previous frame's motion?\\
    \textbf{10:} The specific action is clearly performed; motion is physically plausible and connects smoothly to the previous frame.
    \textbf{8--9:} Instruction followed but motion feels slightly stiff or magnitude is slightly off.
    \textbf{5--7:} Instruction partially ignored, OR motion is abrupt/unnatural (teleporting objects, impossible physics).
    \textbf{2--4:} Instruction mostly ignored or action contradicts the previous frame.
    \textbf{0--1:} Complete hallucination or irrelevant content. \\
    Focus on: \textit{Adherence} (did the requested event happen?), \textit{Physics} (is movement logical relative to the last frame?). \\

    \textit{Metric 2 --- Image Sequence Consistency Score (ISCS), 0--10:}
    Does the image strictly preserve identity, style, and background details from the reference images?\\
    \textbf{10:} Indistinguishable consistency; character, object, and background details are identical to the reference.
    \textbf{8--9:} Minor lighting or shading differences; subject identity clearly preserved.
    \textbf{5--7:} Noticeable ``Identity Drift'' (face looks different, clothing changes) or background objects shift/disappear.
    \textbf{2--4:} Major inconsistencies; looks like a different character or location.
    \textbf{0--1:} Completely different style or content. \\
    Focus on: \textit{Identity} (penalize heavily for face morphing), \textit{Stability} (penalize for flickering or random artifacts). \\
    Overall Reasoning Guide: State clearly whether score deduction is due to ``Instruction Failure'' (Metric~1) or ``Visual Inconsistency'' (Metric~2).

    \end{tcolorbox}

    \caption{Evaluation prompt for the Temporal task.}
    \label{fig:prompt_temporal}
\end{figure*}
\clearpage